\documentclass[10pt,twocolumn,letterpaper]{article}

\usepackage{cvpr}              

\usepackage{subcaption}
\usepackage{graphicx}
\usepackage{amsmath}
\usepackage{amssymb}
\usepackage{booktabs,multirow}
\usepackage[accsupp]{axessibility}  

\usepackage[pagebackref,breaklinks,colorlinks]{hyperref}

\usepackage[capitalize]{cleveref}
\crefname{section}{Sec.}{Secs.}
\Crefname{section}{Section}{Sections}
\Crefname{table}{Table}{Tables}
\crefname{table}{Tab.}{Tabs.}

\def\confName{CVPR}
\def\confYear{2023}

\begin{document}

\title{NTIRE 2023 Quality Assessment of Video Enhancement Challenge}

               

\author{
 Xiaohong Liu$^{*}$ \and Xiongkuo Min$^{*}$ \and Wei Sun$^{*}$ \and  Yulun Zhang $^{*}$ \and Kai Zhang$^{*}$ \and  Radu Timofte$^{*}$ \and  Guangtao Zhai$^{*}$ \and  Yixuan Gao$^{*}$ \and  Yuqin Cao$^{*}$ \and Tengchuan Kou$^{*}$ \and Yunlong Dong$^{*}$ \and Ziheng Jia
 \thanks{The organizers of the NTIRE 2023 Quality Assessment of Video Enhancement Challenge.}
\and Yilin Li$^{\dag}$\and Wei Wu$^{\dag}$ \and Shuming Hu$^{\dag}$\and  Sibin Deng$^{\dag}$\and Pengxiang Xiao$^{\dag}$\and Ying Chen$^{\dag}$\and  Kai Li$^{\dag}$
\and Kai Zhao$^{\dag}$\and Kun Yuan$^{\dag}$\and Ming
Sun$^{\dag}$
\and Heng Cong$^{\dag}$\and Hao Wang$^{\dag}$\and Lingzhi Fu$^{\dag}$\and Yusheng Zhang$^{\dag}$\and Rongyu Zhang$^{\dag}$
\and Hang Shi$^{\dag}$\and Qihang Xu$^{\dag}$\and Longan Xiao$^{\dag}$
\and Zhiliang Ma$^{\dag}$
\and Mirko Agarla$^{\dag}$\and Luigi
Celona$^{\dag}$\and Claudio Rota$^{\dag}$\and Raimondo Schettini$^{\dag}$
\and Zhiwei Huang$^{\dag}$ \and Ya’nan
Li$^{\dag}$ \and Xiaotao Wang$^{\dag}$ \and Lei Lei$^{\dag}$ 
\and Hongye Liu$^{\dag}$\and Wei Hong$^{\dag}$
\and Ironhead Chuang$^{\dag}$ \and Allen Lin$^{\dag}$ \and Drake Guan$^{\dag}$ \and Iris Chen$^{\dag}$ \and Kae Lou$^{\dag}$ \and Willy
Huang$^{\dag}$ \and Yachun Tasi$^{\dag}$ \and Yvonne Kao$^{\dag}$ 
\and Haotian Fan$^{\dag}$\and Fangyuan
Kong$^{\dag}$
\and Shiqi Zhou$^{\dag}$ \and Hao Liu$^{\dag}$ 
\and Yu Lai$^{\dag}$ \and Shanshan Chen$^{\dag}$ 
\and Wenqi Wang$^{\dag}$ 
\and Haoning Wu$^{\dag}$ \and Chaofeng Chen$^{\dag}$ 
\and Chunzheng Zhu$^{\dag}$\and Zekun Guo$^{\dag}$
\and Shiling Zhao$^{\dag}$ \and Haibing Yin$^{\dag}$ \and Hongkui
Wang$^{\dag}$ 
\and Hanene Brachemi Meftah$^{\dag}$ \and Sid Ahmed Fezza$^{\dag}$ \and Wassim Hamidouche$^{\dag}$ \and Olivier Déforges$^{\dag}$ 
\and Tengfei Shi$^{\dag}$ 
\and Azadeh Mansouri$^{\dag}$\and Hossein
Motamednia$^{\dag}$\and Amir Hossein Bakhtiari$^{\dag}$\and Ahmad Mahmoudi Aznaveh
 \thanks{The valid participating teams of the NTIRE 2023 Quality Assessment of Video Enhancement Challenge.\\
 The NTIRE 2023 website: \url{https://cvlai.net/ntire/2023/}}
}
\maketitle

\begin{abstract}
This paper reports on the NTIRE 2023 Quality Assessment of Video Enhancement Challenge, which will be held in conjunction with the New Trends in Image Restoration and Enhancement Workshop (NTIRE) at CVPR 2023. This challenge is to address a major challenge in the field of video processing, namely, video quality assessment (VQA) for enhanced videos.
The challenge uses the VQA Dataset for Perceptual Video Enhancement (VDPVE), which has a total of 1211 enhanced videos, including 600 videos with color, brightness, and contrast enhancements, 310 videos with deblurring, and 301 deshaked videos.
The challenge has a total of 167 registered participants. 61 participating teams submitted their prediction results during the development phase, with a total of 3168 submissions. A total of 176 submissions were submitted by 37 participating teams during the final testing phase. Finally, 19 participating teams submitted their models and fact sheets, and detailed the methods they used. Some methods have achieved better results than baseline methods, and the winning methods have demonstrated superior prediction performance.
\end{abstract}

\section{Introduction}
\label{sec:intro}
The importance of video quality assessment (VQA) in the field of video processing is self-evident. It can guide the development of video processing algorithms such as video capture, enhancement, transmission, and display. Therefore, VQA methods have been widely used to evaluate the quality of various videos, such as user-generated content (UGC) videos \cite{sun2022deep}, high dynamic range (HDR) videos \cite{narwaria2015study}, tone-mapped videos \cite{yeganeh2016objective}, compressed videos \cite{liu2018end}, and so on.
Recently, after capturing a large number of videos, people would like to first enhance certain attributes of the videos, such as contrast, brightness, and color, and then upload these enhanced videos to social medias.
Therefore, many video enhancement methods have been proposed \cite{zhang2021learning,haris2020space,chan2022basicvsr++,shi2021video,liu2018robust,shi2022video,liu2020end}. However, the quality levels of these videos processed by various video enhancement methods are different, and evaluating the quality of these enhanced videos is not easy.
Therefore, it is very important to propose an efficient VQA method to accurately predict the quality of enhanced videos.

This NTIRE 2023 Quality Assessment of Video Enhancement Challenge aims to promote the development of the VQA methods for enhanced videos to guide the improvement and enhancement of the performance of video enhancement methods, thereby improving the viewing experience of videos \cite{liu2021exploit,shi2021learning}. We use the VQA Dataset
for Perceptual Video Enhancement (VDPVE) \cite{gao2023vdpve} for this challenge.
This dataset has 1211 videos with different enhancements, which can be divided into three sub-datasets: the first sub-dataset has 600 videos with color, brightness, and contrast enhancement; the second sub-dataset has 310 videos with deblurring; and the third sub-dataset has 301 deshaked videos. Each enhanced video in the VDPVE has 20 subjective opinion scores.

This is the first time that a quality assessment of video enhancement challenge is held at the NTIRE workshop.
The challenge has a total of 167 registered participants. 61 participating teams submitted their prediction results during the development phase, with a total of 3168 submissions. A total of 176 prediction results were submitted by 37 participating teams during the final testing phase. Finally, 19 valid participating teams submitted their final models and fact sheets. They have provided detailed introductions to their VQA methods for enhanced videos.
We provide the detailed results of the challenge in Secion \ref{Challenge Results} and Secion \ref{Challenge Methods}.
We hope that this challenge can promote the development of VQA methods for video enhancement.

 This challenge is one of the NTIRE 2023 Workshop~\footnote{\url{https://cvlai.net/ntire/2023/}} series of challenges on: night photography rendering~\cite{shutova2023ntire_night}, HR depth from images of specular and transparent surfaces~\cite{zama2023ntire_depth}, image denoising~\cite{li2023ntire_dn50}, video colorization~\cite{kang2023ntire_vc}, shadow removal~\cite{vasluianu2023ntire_isr}, quality assessment of video enhancement~\cite{liu2023ntire}, stereo super-resolution~\cite{wang2023ntire_ssr}, light field image super-resolution~\cite{wang2023ntire_lfsr}, image super-resolution ($\times4$)~\cite{zhang2023ntire}, 360° omnidirectional image and video super-resolution~\cite{cao2023ntire}, lens-to-lens bokeh effect transformation~\cite{conde2023ntire_bokeh}, real-time 4K super-resolution~\cite{conde2023ntire_rtsr}, HR nonhomogenous dehazing~\cite{ancuti2023ntire}, efficient super-resolution~\cite{li2023ntire_esr}.

\section{Related Work}
\subsection{VQA dataset}
The successful construction of VQA datasets is the foundation for proposing effective VQA models.
The first successful VQA dataset is the LIVE Video Quality Database \cite{seshadrinathan2010study}, which has 160 videos with compression and transmission distortions.
IVP \cite{zhang2011ivp} provides 138 videos with compression and transmission distortions.
MCL-V \cite{lin2015mcl} contains 96 distorted videos with two typical video distortion
types: compression and compression followed by scaling.
MCL-JCV \cite{wang2016mcl} is an
H.264/AVC coded video quality dataset consisting of 30 video clips
of a wide content variety.
In recent years, due to the explosion in the number of UGC videos, many researchers have created VQA datasets for UGC videos.
For example, Nuutinen \emph{et al}. introduced the CVD2014 \cite{nuutinen2016cvd2014}, which consists of 234 videos captured by 78 different video capture devices.
The authors in \cite{hosu2017konstanz} constructed one of the most famous VQA datasets for UGC videos, called KoNViD-1k. This dataset has 1200 UGC videos with authentic distortions.
The other two popular VQA datasets for UGC videos are the LIVE-VQC \cite{sinno2018large} and the YouTube-UGC \cite{wang2019youtube}, with 585 and 1380 UGC videos, respectively. 
Besides, we provided a VQA dataset for video enhancement called the VDPVE \cite{gao2023vdpve}. The videos in this dataset were processed by various video enhancement methods, including 600 videos with
color, brightness, and contrast enhancements, 310 videos
with deblurring, and 301 deshaked videos. This dataset is used to test the performance of methods proposed by different participating teams in this challenge.
\subsection{VQA model}
The traditional VQA methods are handcrafted feature-based models. This kind of methods first calculate the quality of each frame of a video by extracting quality features, and then obtain the video quality score \cite{min2018blind,zhai2020perceptual,gao2022image}. For example,
 V-BLIINDS \cite{saad2014blind} is a spatio-temporal natural scene statistics
(NSS) model, which can quantify motion coherency in video scenes.
TLVQM \cite{korhonen2019two} is based on the idea of calculating features at two levels, that is, first calculating the low complexity features of the entire sequence, and then extracting high complexity features from subsets of representative video frames.
VIDEVAL \cite{tu2021ugc} calculates video quality by extracting abundant spatio-temporal features such as motion, jerkiness, blurriness, noise, blockiness, color, and so on.
RAPIQUE \cite{tu2021rapique} combines the advantages of both quality-aware scene statistics features and semantics-aware deep
convolutional features to calculate video quality.

In addition to traditional VQA methods, deep learning-based VQA methods also attract researchers' attention \cite{gao2022imageACM,cao2023attention,sun2021deep,yi2021attention,lu2022deep,zhang2022surveillance}. For example, VSFA \cite{li2019quality} first extracts semantic features from a pre-trained convolutional neural network (CNN), and then uses a gated recursive unit network to extract the temporal relationship between semantic features of video frames to predict video quality.
BVQA \cite{liu2021spatiotemporal} uses a feature encoder to directly extract spatio-temporal representations from videos to predict video quality.
SimpleVQA \cite{sun2022deep} trains an end-to-end spatial feature extraction network to directly learn quality-aware spatial features from video frames, and extracts motion features to measure temporally related distortions that cannot be modeled by spatial features at the same time to predict video quality.

\section{NTIRE 2023 Quality Assessment of Video Enhancement Challenge}
We organize the NTIRE 2023 Quality Assessment of Video Enhancement Challenge in order to promote the development of objective VQA methods for video enhancement. The main goal of the challenge is to predict the perceptual quality of enhanced videos, which can also promote the development of video enhancement methods. Details about the challenge are as follows:
\subsection{Overview}
The challenge has only one track, that is, the task of predicting the perceptual quality of an enhanced video based on a set of prior examples of videos and their perceptual quality labels. The challenge uses the training, validation, and testing sets as defined in the VDPVE \cite{gao2023vdpve}. As the final result, the participants in the challenge are asked to submit predicted scores for the given testing set.

\subsection{Dataset}
The VDPVE has 1211 videos with different enhancements, which can be divided into three sub-datasets: the first sub-dataset has 600 videos with color, brightness, and contrast enhancements; the second sub-dataset has 310 videos with deblurring; and the third sub-dataset has 301 deshaked videos. The resolution of all videos in the VDPVE is $1280\times720$. The video length is 8s or 10s.

In the first sub-dataset, eight enhancement methods are utilized to enhance the color, brightness, and contrast of 79 videos: ACE \cite{getreuer2012automatic}, AGCCPF \cite{gupta2016minimum}, BPHEME \cite{wang2005brightness}, MBLLEN \cite{lv2018mbllen}, SGZSL \cite{zheng2022semantic}, DCC-Net \cite{zhang2022deep}, and two commercial software: CapCut and Adobe Premiere Pro. In the second sub-dataset, we utilize five enhancement methods to deblur the $62$ blurred videos, including ESTRNN \cite{zhong2020efficient}, DeblurGANv2 \cite{kupyn2019deblurgan}, FGST \cite{lin2022flow}, BasicVSR++ \cite{chan2022basicvsr++}, and Adobe Premiere Pro. In the third sub-dataset, seven enhancement methods are utilized to stabilize $43$ videos, including GlobalFlowNet \cite{james2023globalflownet}, DIFRINT \cite{choi2020deep}, PWStableNet \cite{zhao2020pwstablenet}, Yu \cite{yu2020learning}, CapCut (most stable mode), CapCut (minimum cropping mode), and Adobe Premiere Pro. 

We invited $21$ subjects ($20$ valid subjects) to rate all enhanced videos in the VDPVE. After normalizing and averaging the subjective opinion scores, the mean opinion score (MOS) of each video can be obtained. Furthermore, we randomly split the enhanced videos in the VDPVE into a training set, a validation set, and a testing set according to the ratio of $7:1:2$. The enhanced videos generated from the same original video are divided into the same set. The numbers of enhanced videos in the training set, validation set, and testing set are $839$, $119$, and $253$, respectively.
\subsection{Evaluation protocol}
In the challenge, the main scores are utilized to determine the rankings of participating teams. We ignore the sign and calculate the average of Spearman rank-order correlation coefficient (SRCC) and Person linear correlation coefficient (PLCC) as the main score:
\begin{equation}
    \mathrm{Main\;Score} = (|\mathrm{SRCC}| + |\mathrm{PLCC}|)/2.
\end{equation}
SRCC measures the prediction monotonicity, while PLCC measures the prediction accuracy. Better VQA methods should have larger SRCC and PLCC values. Before calculating PLCC index, we perform the third-order polynomial nonlinear regression. By combining SRCC and PLCC, the main scores can comprehensively measure the performance of participating methods.
\subsection{Challenge phases}
The whole challenge consists of two phases: the developing phase and the testing phase. In the developing phase, the participants can access to the enhanced videos of the training set and the corresponding MOSs. Participants can be familiar with dataset structure and develop their VQA methods. We also release the enhanced videos of the validation set without corresponding MOSs. Participants can utilize their VQA methods to predict the quality scores of the validation set and upload the results to the server. The participants can receive immediate feedback and analyze the effectiveness of their methods on the validation set. The validation leaderboard is available. In the testing phase, the participants can access to the enhanced videos of the testing set without MOSs and upload the final predicted scores of the testing set before the challenge deadline. Each participating team needs to submit a source code/executable and a fact sheet, which is a detailed description file of the proposed method and the corresponding team information. The final results are then sent to the participants.

\section{Challenge Results}
\label{Challenge Results}
\begin{table*}[!t]
    \centering
    \caption{Quantitative results for the NTIRE 2023 Quality Assessment of Video Enhancement Challenge.}
    \resizebox{0.8\textwidth}{!}{
    \begin{tabular}{c|c|c|ccc}
    \toprule
    Rank & Team & Leader & Main Score & SRCC & PLCC \\
    \midrule
    1 & TB-VQA & Yilin Li & 0.8576 & 0.8493 & 0.8659\\
    2 & QuoVadis & Kai Zhao & 0.8396 & 0.8408 & 0.8383 \\
    3 & OPDAI & Heng Cong & 0.8289 & 0.8261 & 0.8317 \\
    4 & TIAT & Hang Shi & 0.8199 & 0.8163 & 0.8236 \\
    5 & VCCIP & Zhiliang Ma & 0.7994 & 0.7962 & 0.8026 \\
    6 & IVL & Mirko Agarla & 0.7859 & 0.7896 & 0.7822 \\
    7 & HXHHXH & Zhiwei Huang & 0.7850 & 0.7879 & 0.7821 \\
    8 & fmgtv & Hongye Liu & 0.7727 & 0.7756 & 0.7698 \\
    9 & KK-ARC & Ironhead Chuang & 0.7635 & 0.7663 & 0.7607 \\
    10 & DTVQA & Haotian Fan & 0.7325 & 0.7357 & 0.7294 \\ 
    11 & sqiyx & Shiqi Zhou & 0.7302 & 0.7246 & 0.7358 \\
    12 & 402Lab & Yu Lai & 0.7136 & 0.7150 & 0.7123 \\
    13 & one\_for\_all & Wenqi Wang & 0.6990 & 0.7087 & 0.6893\\
    14 & NTU-SLab & Haoning Wu & 0.6972 & 0.7019 & 0.6924 \\
    15 & HNU-LIMMC & Chunzheng Zhu & 0.6923 & 0.6975 & 0.6872 \\
    16 & Drealitym & Shiling Zhao & 0.6863 & 0.6900 & 0.6826 \\
    17 & LION\_Vaader & Hanene Brachemi Meftah & 0.6596 & 0.6674 & 0.6518 \\
    18 & Caption Timor & Tengfei Shi & 0.6499 & 0.6524 & 0.6475 \\
    19 & IVP-LAB & Azadeh Mansouri & 0.5851 & 0.5887 & 0.5814 \\
    \midrule
    \multirow{8}{*}{Baseline} & \multicolumn{2}{c|}{FastVQA} & 0.7330 & 0.7350 & 0.7310 \\
    ~ & \multicolumn{2}{c|}{BVQA} & 0.6835 & 0.6995 & 0.6674\\
    ~ & \multicolumn{2}{c|}{SimpleVQA} & 0.6347 & 0.6340 & 0.6354\\
    ~ & \multicolumn{2}{c|}{VSFA} & 0.5648 & 0.5871 & 0.5424\\
    ~ & \multicolumn{2}{c|}{V-BLIINDS} & 0.5578 & 0.5652 & 0.5503\\
    ~ & \multicolumn{2}{c|}{TLVQM} & 0.5492 & 0.5474 & 0.5509 \\
    ~ & \multicolumn{2}{c|}{RAPIQUE} & 0.5414 & 0.5434 & 0.5393 \\
    ~ & \multicolumn{2}{c|}{VIDEVAL} & 0.4865 & 0.5005 & 0.4724 \\
    \bottomrule
    \end{tabular}
    }
    \label{tab:Quantitative results}
\end{table*}

\begin{figure*}[ht]
\centering
\begin{subfigure}{0.19\textwidth}
    \includegraphics[width=\textwidth]{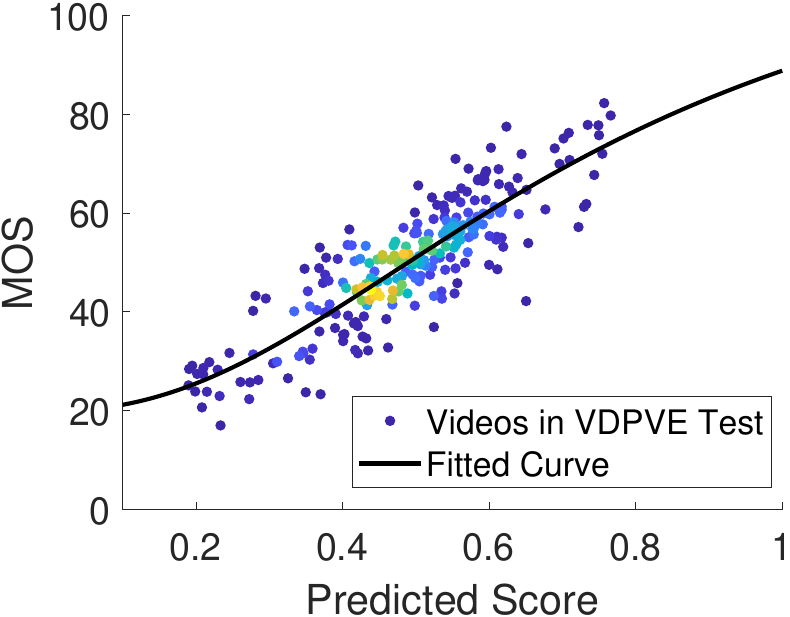}
    \caption{TB-VQA}
    \label{fig:first}
\end{subfigure}
\begin{subfigure}{0.19\textwidth}
    \includegraphics[width=\textwidth]{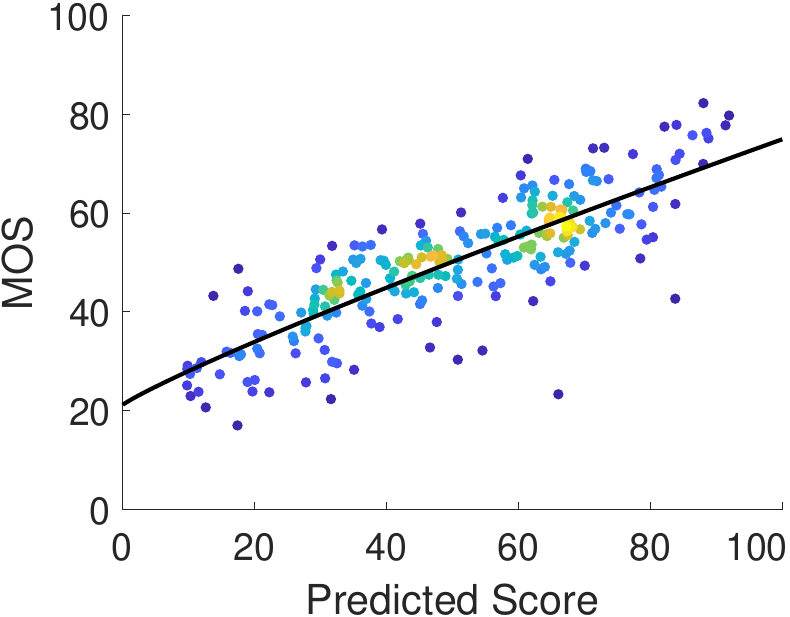}
    \caption{QuoVadis}
    \label{fig:first}
\end{subfigure}
\begin{subfigure}{0.19\textwidth}
    \includegraphics[width=\textwidth]{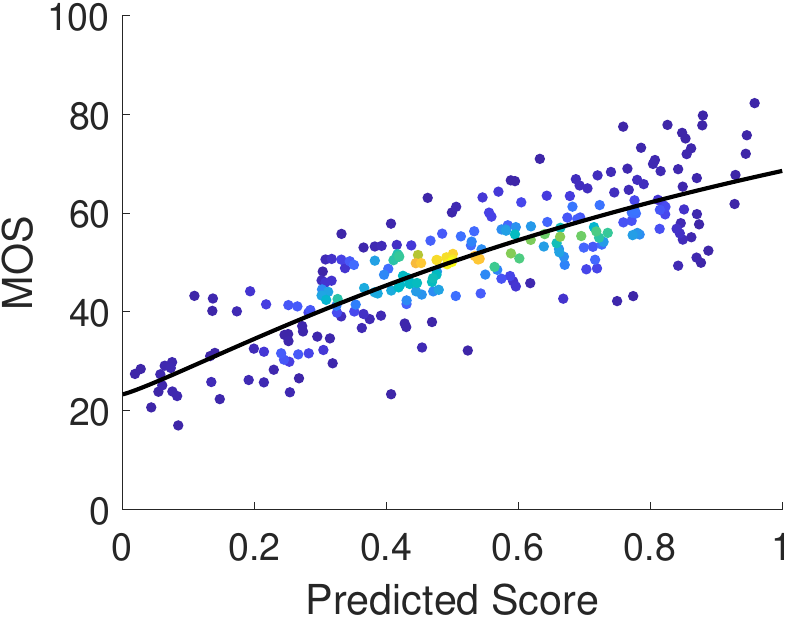}
    \caption{OPDAI}
    \label{fig:first}
\end{subfigure}
\begin{subfigure}{0.19\textwidth}
    \includegraphics[width=\textwidth]{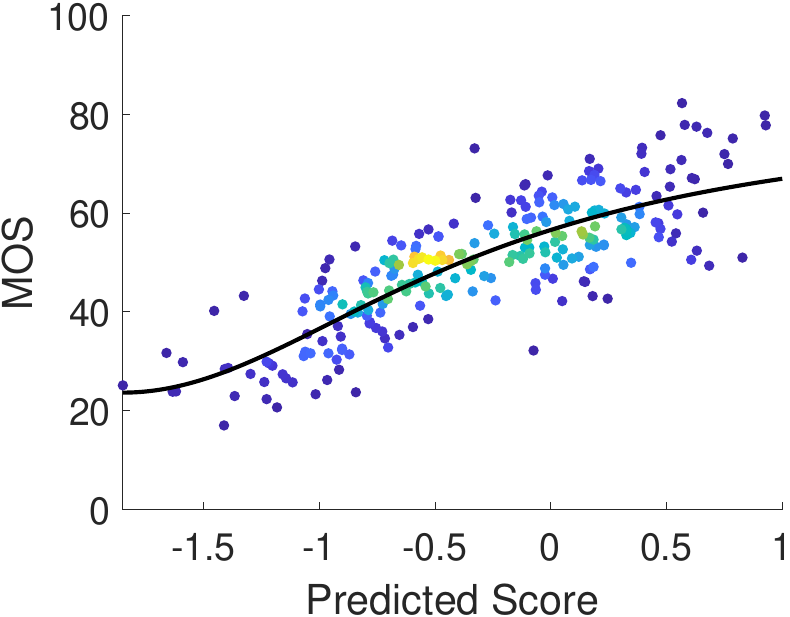}
    \caption{TIAT}
    \label{fig:first}
\end{subfigure}
\begin{subfigure}{0.19\textwidth}
    \includegraphics[width=\textwidth]{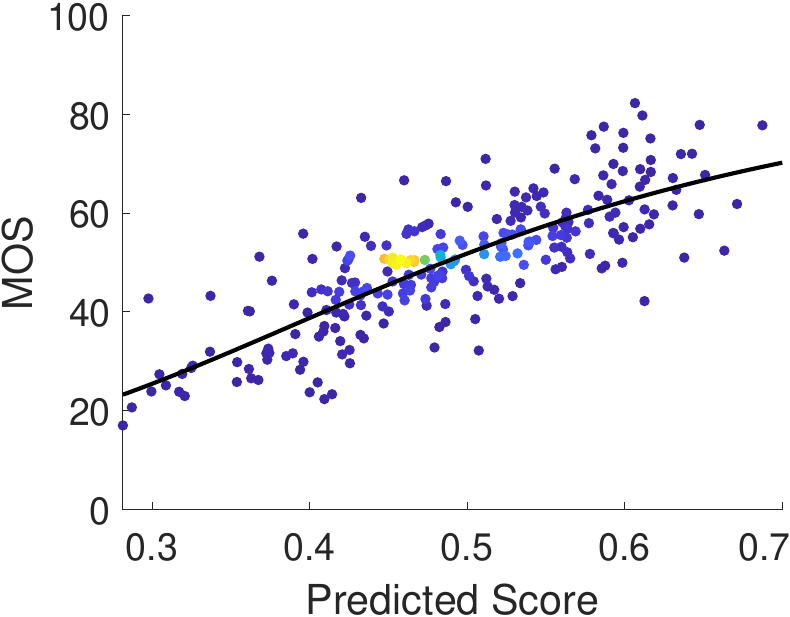}
    \caption{VCCIP}
    \label{fig:first}
\end{subfigure}
\begin{subfigure}{0.19\textwidth}
    \includegraphics[width=\textwidth]{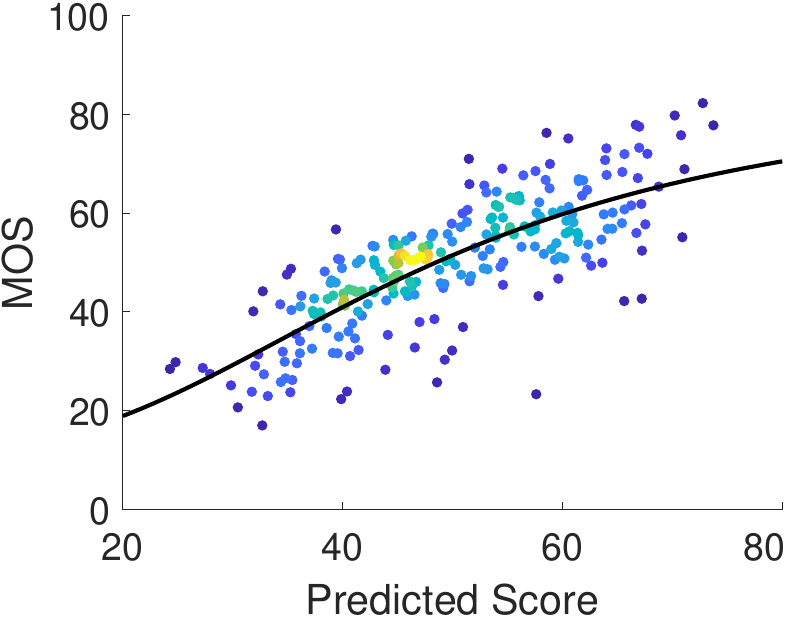}
    \caption{IVL}
    \label{fig:first}
\end{subfigure}
\begin{subfigure}{0.19\textwidth}
    \includegraphics[width=\textwidth]{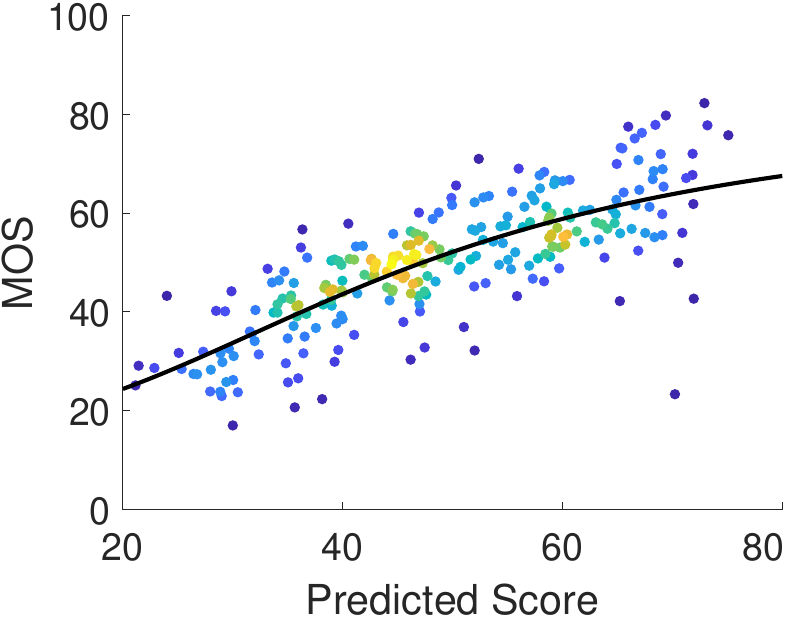}
    \caption{HXHHXH}
    \label{fig:first}
\end{subfigure}
\begin{subfigure}{0.19\textwidth}
    \includegraphics[width=\textwidth]{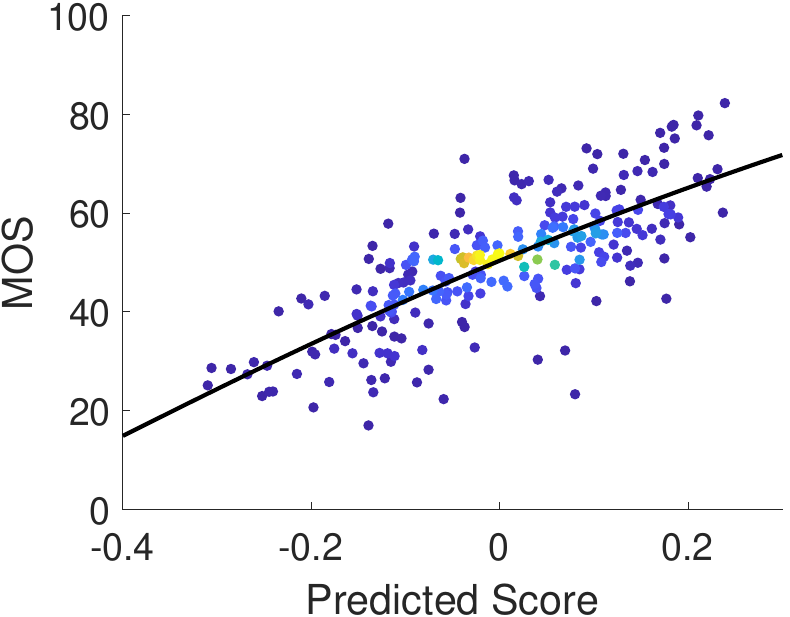}
    \caption{fmgtv}
    \label{fig:first}
\end{subfigure}
\begin{subfigure}{0.19\textwidth}
    \includegraphics[width=\textwidth]{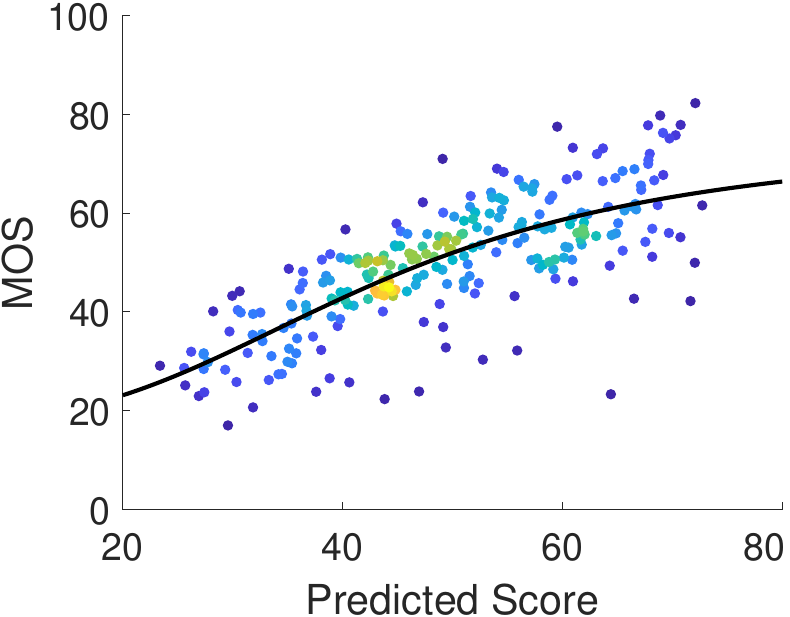}
    \caption{KK-ARC}
    \label{fig:first}
\end{subfigure}
\begin{subfigure}{0.19\textwidth}
    \includegraphics[width=\textwidth]{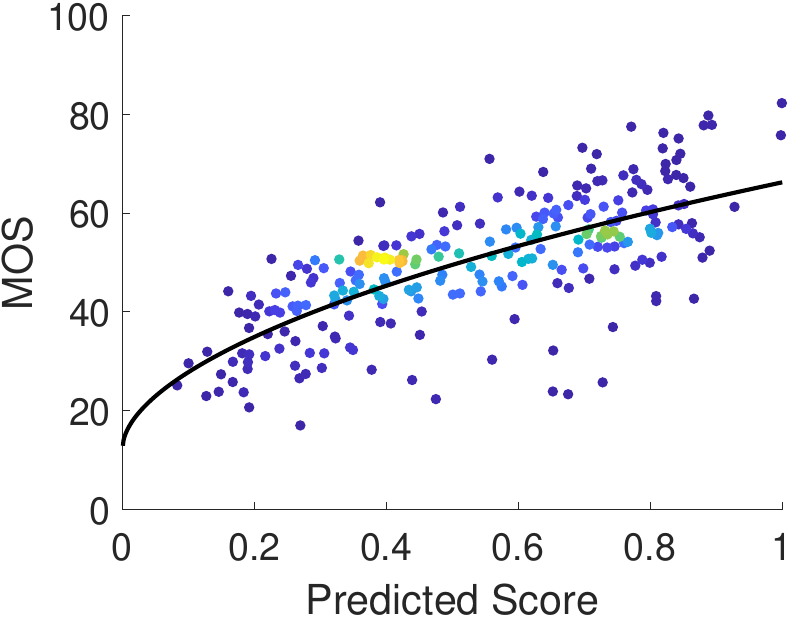}
    \caption{DTVQA}
    \label{fig:first}
\end{subfigure}
\begin{subfigure}{0.19\textwidth}
    \includegraphics[width=\textwidth]{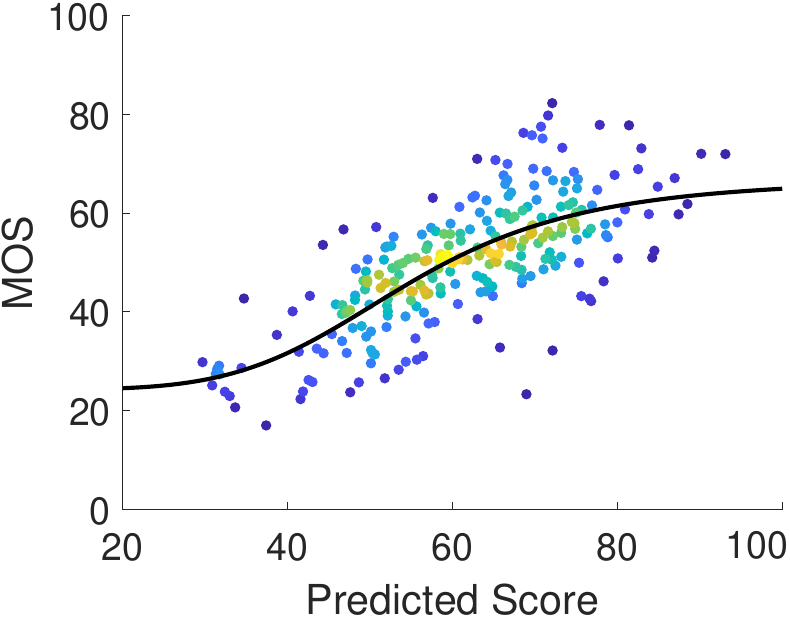}
    \caption{sqiyx}
    \label{fig:first}
\end{subfigure}
\begin{subfigure}{0.19\textwidth}
    \includegraphics[width=\textwidth]{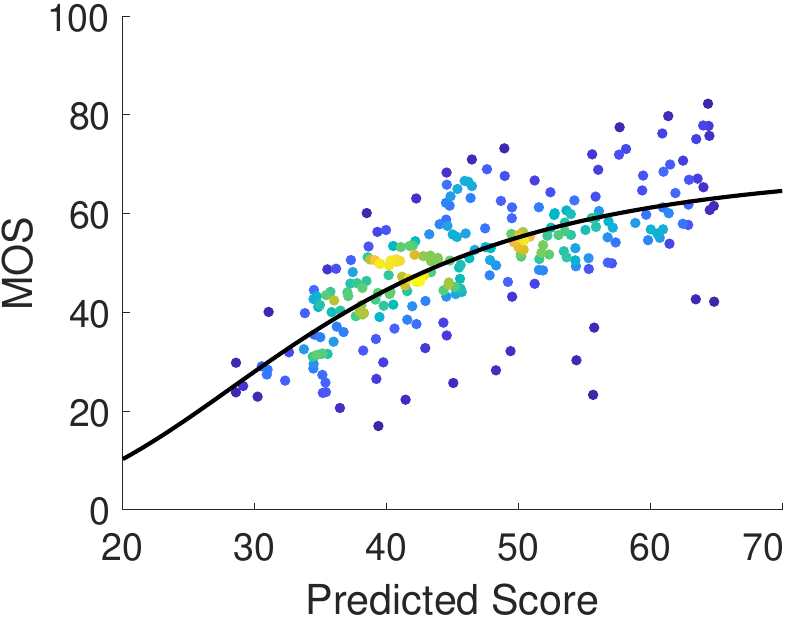}
    \caption{402Lab}
    \label{fig:first}
\end{subfigure}
\begin{subfigure}{0.19\textwidth}
    \includegraphics[width=\textwidth]{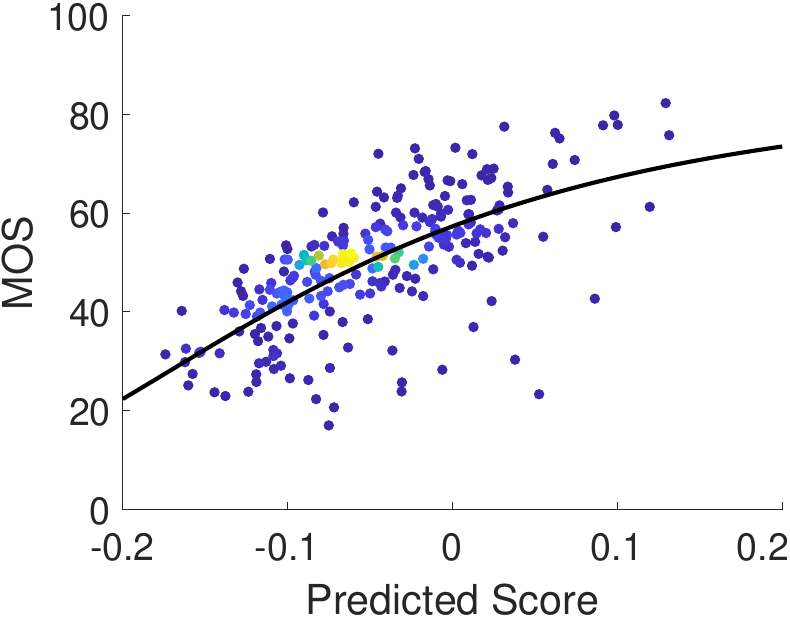}
    \caption{one\_for\_all}
    \label{fig:first}
\end{subfigure}
\begin{subfigure}{0.19\textwidth}
    \includegraphics[width=\textwidth]{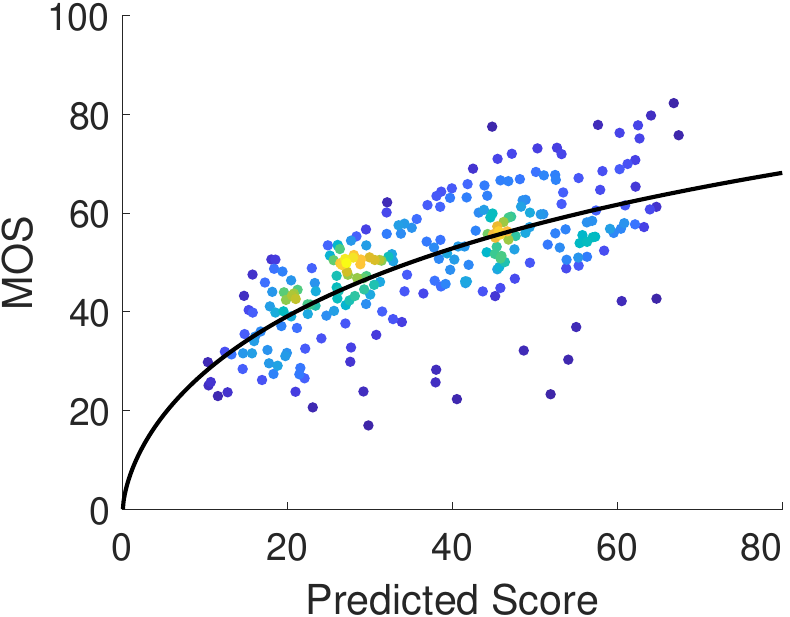}
    \caption{NTU-SLab}
    \label{fig:first}
\end{subfigure}
\begin{subfigure}{0.19\textwidth}
    \includegraphics[width=\textwidth]{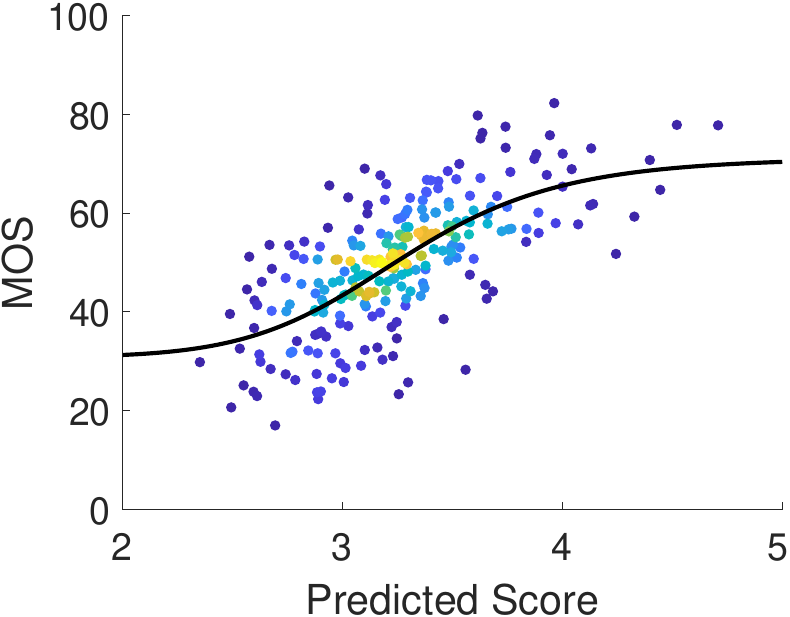}
    \caption{HNU-LIMMC}
    \label{fig:first}
\end{subfigure}
\begin{subfigure}{0.19\textwidth}
    \includegraphics[width=\textwidth]{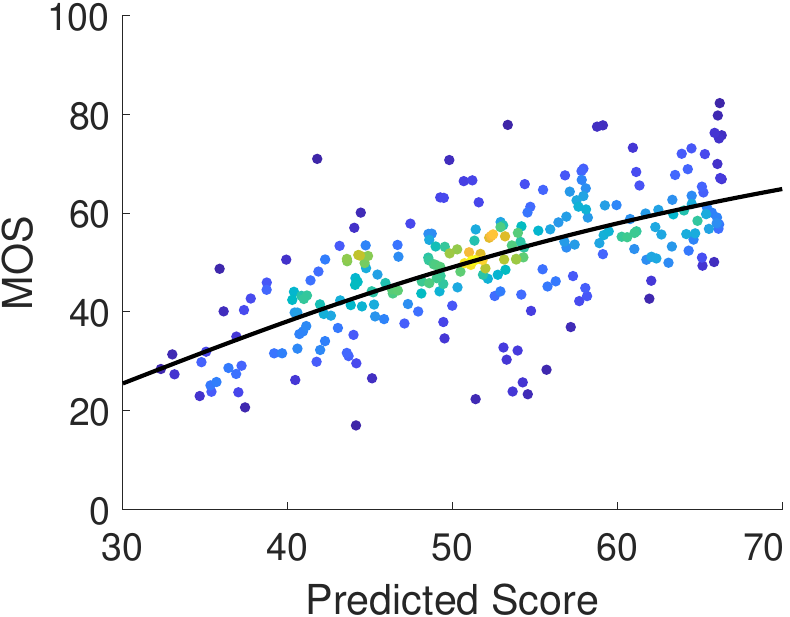}
    \caption{Drealitym}
    \label{fig:first}
\end{subfigure}
\begin{subfigure}{0.19\textwidth}
    \includegraphics[width=\textwidth]{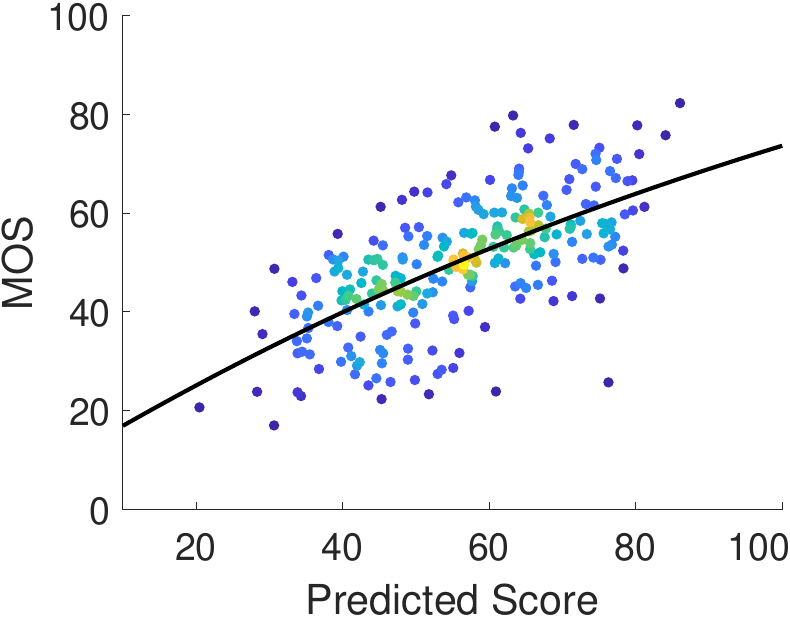}
    \caption{LION\_Vaader}
    \label{fig:first}
\end{subfigure}
\begin{subfigure}{0.19\textwidth}
    \includegraphics[width=\textwidth]{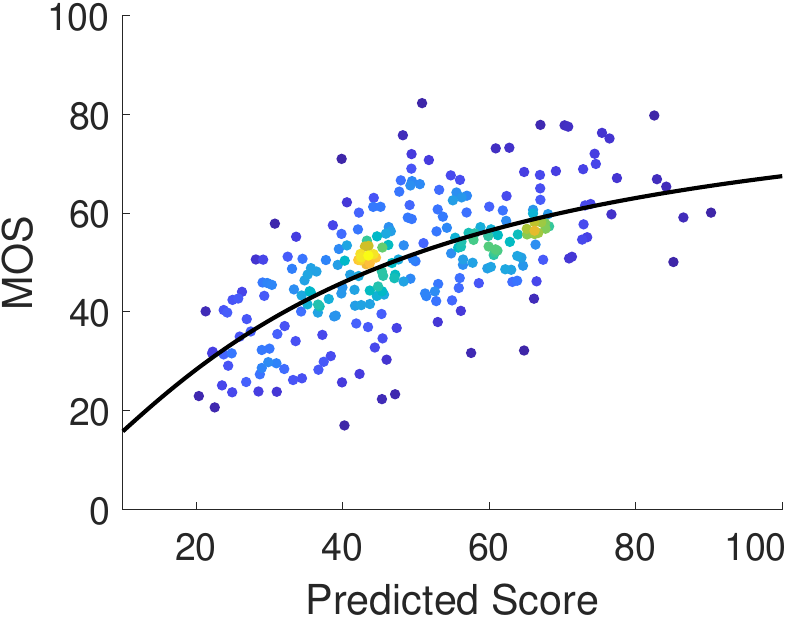}
    \caption{Caption Timor}
    \label{fig:first}
\end{subfigure}
\begin{subfigure}{0.19\textwidth}
    \includegraphics[width=\textwidth]{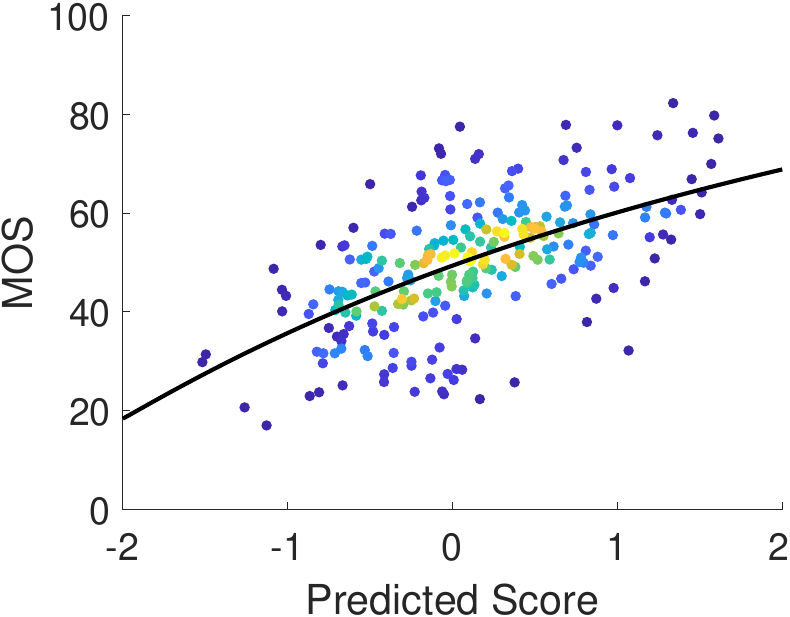}
    \caption{IVP-LAB}
    \label{fig:first}
\end{subfigure}
\caption{Scatter plots of the predicted scores vs. MOSs. The curves are obtained by a four-order polynomial
nonlinear fitting}
\label{fig:Scatter plots}
\end{figure*}

A total of 37 teams participated in the testing phase of NTIRE 2023 Quality Assessment of Video Enhancement Challenge, and 19 teams submitted their final codes/executables and fact sheets. Table \ref{tab:Quantitative results} summarizes the main results and important information of the 19 valid teams. The methods of these teams are briefly introduced in Section \ref{Challenge Methods} and the team members are listed in Appendix \ref{sec:apd:track1team}.

\subsection{Baselines}
We compare the performance of submitted methods with several state-of-the-art NR VQA methods on the testing set, including V-BLIINDS \cite{saad2014blind}, TLVQM \cite{korhonen2019two}, VIDEVAL \cite{tu2021ugc}, RAPIQUE \cite{tu2021rapique}, FastVQA \cite{wu2022fast}, VSFA \cite{li2019quality}, BVQA \cite{liu2021spatiotemporal}, and SimpleVQA \cite{sun2022deep}. V-BLIINDS \cite{saad2014blind}, TLVQM \cite{korhonen2019two}, VIDEVAL \cite{tu2021ugc}, and RAPIQUE \cite{tu2021rapique} are the handcrafted feature-based VQA models. FastVQA \cite{tu2021rapique}, VSFA \cite{li2019quality}, BVQA \cite{liu2021spatiotemporal} and SimpleVQA \cite{sun2022deep} are deep learning-based VQA models. VSFA \cite{li2019quality}, BVQA \cite{liu2021spatiotemporal}, and SimpleVQA \cite{sun2022deep} utilize CNN models as the network backbone, while FastVQA \cite{tu2021rapique} utilizes the transformer as the network backbone.
\subsection{Discussion}
The main results of 19 teams' methods and the baseline methods are shown in Table \ref{tab:Quantitative results}, It can be seen that most of existing NR VQA methods are not ideal on VDPVE testing set, while the submitted methods have basically achieved good results. It means that these methods are closer to human visual perception when used to evaluate enhanced videos. $9$ teams achieve relatively better performance than FastVQA, which has good performance on the in-the-wild VQA task. Furthermore, the main scores of $4$ teams exceed $0.8$. The championship team achieves the SRCC score of $0.8576$ and the PLCC score of $0.8396$. Figure \ref{fig:Scatter plots} shows scatter plots of predicted scores versus MOSs for the 19 teams' methods on VDPVE testing set. The curves shown in Figure \ref{fig:Scatter plots} are obtained by a four-order polynomial nonlinear fitting. We can observe that the predicted scores obtained by the top team methods have higher correlations with the MOSs. They can not only meet the need to predict quality scores for enhanced videos but also contribute to improving the performance of video enhancement methods.

\section{Challenge Methods}
\label{Challenge Methods}
\subsection{TB-VQA}
Team TB-VQA is the final winner of the challenge. They propose a NR VQA method~\cite{wu2023video} based on Swin Transformer with improved spatio-temporal feature fusion and data sample augmentation strategy. The network is developed on top of SimpleVQA \cite{sun2022deep} and is composed of two key components: the spatial feature extraction module and spatio-temporal feature fusion module. In the spatial feature extraction module, inspired by its strong modeling capabilities and representative performance of visual priors including hierarchy, locality, and translation invariance, they exploit Swin Transformer V2 \cite{liu2022swin}, pre-trained on ImageNet-1K \cite{deng2009imagenet}, as the backbone network to extract spatial features. Specially, they adopt the features extracted from the last two Transformer blocks to take advantage of deep semantic information in video quality representation. In the spatio-temporal feature fusion module, they introduce an $1\times1$ convolutional layer, which deepens the spatial features extracted from the intermediate stages of the pre-trained network, to mitigate the gap between shallow and deep features. The semantic features, the deepened features, and the temporal features (originally from the motion feature extraction module in \cite{sun2022deep}) are flattened and fused as the final features for video quality prediction. To further enhance the robustness of the model, a data augmentation strategy is performed to increase the number of video frames in the training phase. Specially, they first train their model on the LSVQ \cite{ying2021patch} dataset. Then, they build a dataset similar to the VDPVE training set and further train the model on the dataset they built, and then fine-tune it on the VDPVE training set. Particularly, multiple temporal frames are randomly sampled from each segment of videos, which efficiently increases the volume of the training dataset. As a comparison, only one frame per segment with fixed sampling order is utilized as a training sample in \cite{sun2022deep}. 

There are $29.77$ million trainable parameters in the model. In the training phase, the input frames are resized to $320\times320$ and randomly cropped with a resolution of $256\times256$. Batch size is set as 16 and Adam optimizer with $\beta_1=0.9, \beta_2=0.999$ is utilized for optimization. The learning rate is initialized as $10^{-5}$ and decayed by $\gamma=0.95$ every two epochs. During testing, the input frames are resized to $320\times320$, and the ``torchvision.transforms.TenCrop" function is used to crop $10$ image patches with a resolution of $256\times256$, which are located at the four corners and the center, respectively, as well as the horizontally flipped version of the previous crops.

\subsection{QuoVadis}
\begin{figure*}[ht]
\centering
\includegraphics[width=\textwidth]{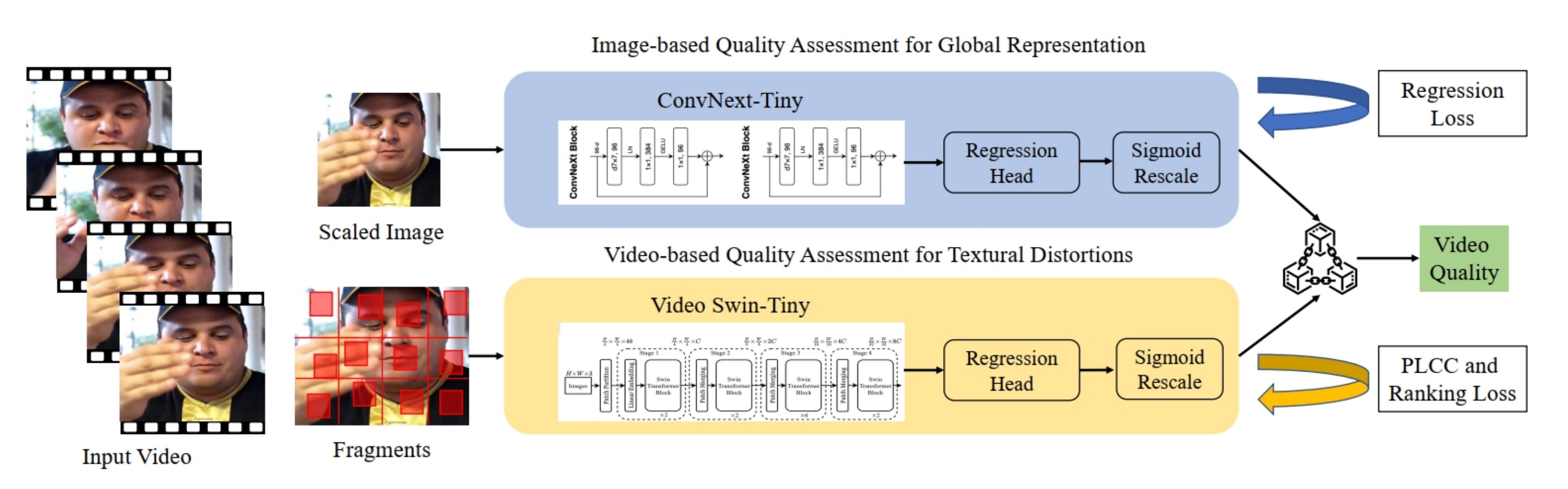}
\caption{The overview of QuoVadis team proposed dual-branch VQA network for enhanced videos.}
\label{figure2}
\end{figure*}

Team QuoVadis wins second place in the challenge. They propose a dual-branch VQA network for enhanced videos~\cite{zhao2023zoom}. As is shown in Figure \ref{figure2}, the overall architecture consists of two parts: the image-based network and the video-based network. The image-based network receives single images as input and generates quality prediction in a global view, while the video-based one obtains shuffled fragments and predicts prediction focusing on textural distortions. 

Specially, the architecture used for the image-based network is ConvNext-Tiny \cite{liu2022convnet}, following a regression head. To analyze the overall quality of the video, they uniformly sample two frames per second and input them into the network. For each frame, they scale the shorter side to $512$ and maintain the same scaling ratio for the longer side. They then crop a $320\times320$ region from the center as the network input to obtain the global quality information of the image. They further attach a patch-weighted quality prediction head as \cite{yang2022maniqa}. And the final prediction is generated by the multiplication of each patch’s score and weight. The outputs generated by all frames are averaged as the whole quality of the video. During training, the network is optimized using a smooth $\mathcal{L}_1$ regression loss. Given a video $X$ and corresponding sampled frames $\{x_1, x_2, ..., x_n\}$, the optimization objective can be written as:
$$min\;\mathcal{L}_{reg} = min\; \frac{1}{n}\sum_{i = 1} ^n||\mathcal{F}(x_i) - y||_1,$$
where $\mathcal{F}(\cdot)$ is the mapping function of the network, and $y$ is the labeled MOS for the video.

Unlike the above image-based network, the video-based network receives video clips as input. Following \cite{wu2022fast}, each clip contains 32 frames sampled uniformly. To preserve the original video quality and obtain local texture information which benefits quality assessment, they utilize the sampling strategy of fragments used in FastVQA. The fragments are obtained through uniform grid mini-patch sampling. This method vastly reduces the computational cost by $97.6\%$ compared with computing attention on the whole resolution. The image-based network structure described above can perceive global semantic information. To complement this, they redefine the form of the fragments and further randomly shuffle the positions of the mini-patches in space, allowing the network to pay more attention to low-quality texture information (such as noise, blur, block effects, etc.) and reduce its focus on higher-level semantics. Then the shuffled fragments are sent into an attention network of Video Swin-Tiny \cite{liu2022video}. During training, specifically, a PLCC-induced loss and a ranking-based loss are utilized. Assume there are $m$ videos in the training batch. Given the predicted quality scores $\{y_{1}' , y_{2}', ..., y_{m}'\}$ and the MOS values $\{y_1, y_2, ..., y_m\}$, the PLCC-induced loss is defined as:
$$\mathcal{L}_{plcc} = (1 - \frac{\sum_{i = 1} ^m(y_{i}' - a')(y_i - a)}{\sqrt{\sum_{i = 1} ^m(y_{i}' - a')^2\sum_{i = 1} ^m(y_i - a)^2}})/2,$$
where $a'$ and $a$ are the mean values of $m$ predicted quality scores and MOSs respectively. And the ranking-based loss can be denoted as:
$$\mathcal{L}_{rank} = \frac{1}{m^2}\sum_{i = 1} ^m\sum_{i = 1} ^m max(0, |y_i - y_j| - e(y_i, y_j)\cdot(y_{i}' - y_{j}')),$$
where $e(y_i, y_j)$ is $1$ if $y_i \geq y_j$, else is $-1$ if $y_i \textless y_j$. And the optimization objective can be written as:
$$min\;\mathcal{L}_{plcc} + \beta\cdot\mathcal{L}_{rank},$$
and $\beta$ is the coefficient for balancing. In practice, it is set to $0.3$.

There are $55$ million parameters in the entire model. In ConvNext-Tiny, the batch size is set to 32, the epoch is set to 30, the learning rate is initialized as $10^{-4}$, and AdamW with $10^{-2}$ weight decay is utilized for optimization. In Video Swin-Tiny, the batch size is set to 16, the epoch is set to 30, the learning rate is initialized as $10^{-3}$, and AdamW with $10^{-2}$ weight decay is utilized for optimization. During the testing phase, the video is analyzed using the dual-branch network structure based on both images and videos, and the prediction results from both branches are averaged to obtain the final prediction quality. 

\subsection{OPDAI}
Team OPDAI wins third place in the challenge. They apply image quality assessment into VQA. Specifically, they combine one VQA method and four image quality assessment methods for this competition. The VQA is based on DOVER \cite{wu2022disentangling}. The four image quality assessment models are main model SwinTransfromer \cite{liu2021swin}, main model ConvnextV2 \cite{woo2023convnext}, using SwinTransformer as backbone extracting feature and using stacked transformer to regress quality score, and main model CDCNN \cite{zhang2020cdcnn}, respectively. Then, they bagging 5 models to obtain the final results. The mean absolute error (MAE), mean square error (MSE), norm-in-norm loss and Kullback-Leibler (KL) divergence loss are used for training. It is worth noting that not only the provided NTIRE 2023 dataset, but also part of Youtube-UGC \cite{wang2019youtube} dataset and PIPAL \cite{jinjin2020pipal} are used for pretraining the model. They used a cosine annealing learning rate descent method with warming-up. Minibatch size is set to 60, and the learning rate is initialized as $4\times10^{-5}$. They utilize AdamW optimizer setting $\beta_1=0.9, \beta_2=0.999$.

\subsection{TIAT}
TIAT proposes a deep learning based VQA model. In order to try their best to make use of the information contained in the target video, the spatial information and temporal information in each video are extracted and fused through segmented processing of the target video. The model is based on SimpleVQA \cite{sun2022deep} and they modify the spatial feature extraction module. Specifically, they utilize Swin-b \cite{liu2021swin} network pre-trained on ImageNet \cite{deng2009imagenet} and slowfast r50 \cite{feichtenhofer2019slowfast} network pre-trained on Kinetics \cite{carreira2017quo} as the feature extraction model.

There are totally $87.30$ million parameters in the model. In the training phase, minibatch size is set to 6, the training epoch is set to 40, and the learning rate is initialized as $10^{-5}$. They utilize Adam optimizer setting $\beta_1=0.9, \beta_2=0.999$.

\subsection{VCCIP}
Team VCCIP proposes a VQA network model for channel fusion, which integrates information from different channels of three consecutive frames in time. This network takes into account both spatial and temporal information, which enables better performance. Additionally, they add a subtask to identify the category of enhancement method to optimize and train the network model more effectively.
\begin{figure*}[ht]
\centering
\includegraphics[width=\textwidth]{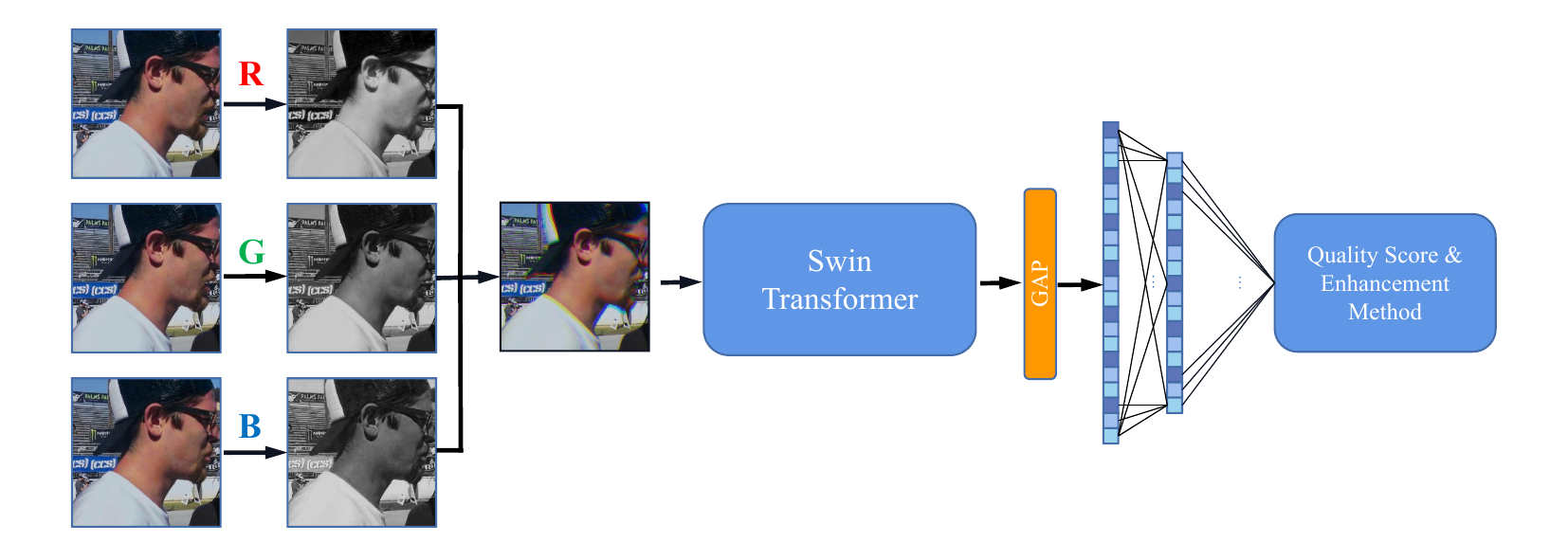}
\caption{The framework design of VCCIP team's method.}
\label{figure3}
\end{figure*}

The architecture of CF-VQA is illustrated in Figure \ref{figure3}, which comprises of a Swin Transformer \cite{liu2021swin} backbone and a quality score regression module. The network structure is kept simple.  To leverage the powerful learning ability of the Transformer structure, this method tends to utilize a pretrained Swin Transformer base as the backbone. After extracting quality perception features through the effective Swin Transformer backbone, a regression model is incorporated to map these features to the quality score. Firstly, the global average feature pooling (GAP) is applied to generate a feature vector with a dimension of $P\times2$, where $P$ represents the number of final feature maps. Then, two fully connected (FC) layers, consisting of 512 neurons and 2 neurons, respectively, are utilized to map the feature vector to the predicted quality score and enhancement method category. Finally, CF-VQA can be trained on VDPVE \cite{gao2023vdpve} using an end-to-end training method with $\mathcal{L}_1$ loss function,
$$\mathcal{L}_1 =  \frac{1}{N}\sum_{i = 1} ^N||q_{score} - q_{label}||_1,$$
where $q_{score}$ and $q_{label}$ denote the predicted score and MOS of the $i$-$th$ training patch, and $N$ represents the total number of training patches.

There are $87.41$ million parameters in the model. Swin Transformer Base Network pre-trained using ImageNet \cite{deng2009imagenet} is utilized as the feature extraction sub-model. Batch size is set to 4, the learning rate is set to $10^{-5}$, and AdamW optimizer is adopted with a weight decay of $5\times10^{-4}$. In addition, they use the cosine decay learning rate with the minimum learning rate of $10^{-7}$, and use linear preheating in first 2 epochs with start learning rate $5\times10^{-7}$. When training on VDPVE, they randomly sample and horizontally flipping with size $384\times384$ pixels from each training image for augmentation. In the test phase, one hundred video clips with $384\times384$ pixels are randomly cropped from each video, and the final quality score of a video is the average score of all clips.

\subsection{IVL}
\begin{figure*}[ht]
\centering
\includegraphics[width=\textwidth]{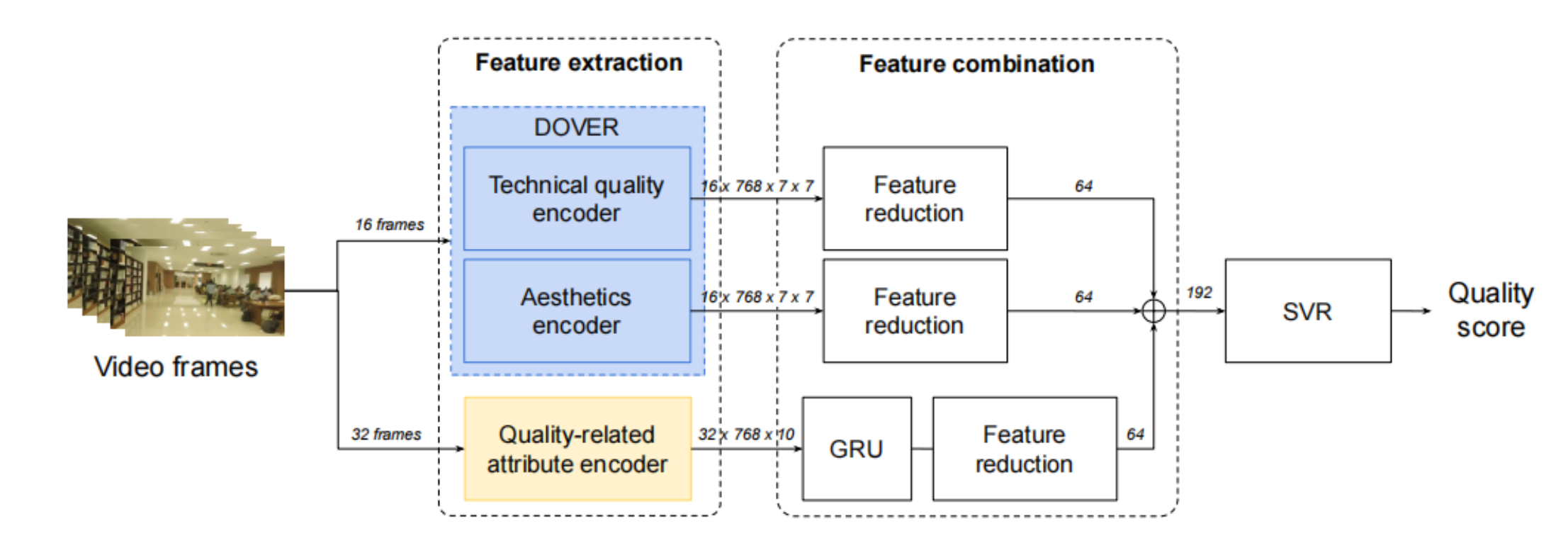}
\caption{The framework design of IVL team's method.}
\label{figure4}
\end{figure*}

Team IVL proposes a VQA method~\cite{agarla2023quality} inspired by DOVER \cite{wu2022disentangling} and the NR-VQA model introduced in \cite{agarla2021efficient}. As depicted in Figure \ref{figure4},  it consists of three components: the feature extractor module includes the technical quality and aesthetic encoders and the quality-related attribute encoder; the feature combination module provides a temporal aggregation part for the frame-level features of the quality-related attribute encoder and reduces the feature vectors obtained by all the encoders to a fixed size; the quality prediction module exploits a support vector regression (SVR) machine for mapping the feature vector into the video quality score.

The feature extraction module contains DOVER to model features capturing information about distortion perception (technical quality) and preferences (aesthetics), and a quality-related attribute encoder to model quality features capturing various quality attributes, such as brightness, contrast and sharpness. The DOVER architecture is modified so that its outputs are feature vectors rather than quality scores. The aesthetic encoder consists of a tiny inflated-ConvNext \cite{liu2022convnet} as backbone while the technical quality encoder exploits a tiny Video Swin Transformer \cite{liu2022video}. The aesthetic encoder has an overall view of the video as it uses a total of 32 equally-spaced frames covering the entire video sequence. The frames processed by this encoder are downscaled to a 480p resolution to increase efficiency. The technical quality encoder divides the video into two parts, and selects 32 frames in the first part and 32 frames in the last one with a stride of 4 to increase video coverage. Here five-crop video fragments are used, \textit{i.e.}, each frame is cropped at the four corners and at the central, and fragments are generated from these crops. Note that each crop has a 480p resolution, thus the fragments can cover about 80\% of the entire frame spatial dimension. Moreover, downscaling is performed using soft pooling \cite{stergiou2021refining} to better preserve video distortions and avoid masking effects. The fragments are processed independently from the others, and the final features are obtained by averaging the features extracted from each crop. The quality-related attribute encoder is similar to the one proposed in \cite{agarla2021efficient}, but the MobileNet-v2 \cite{sandler2018mobilenetv2} backbone is replaced with EfficientNet-v2 \cite{tan2021efficientnetv2} because of its higher capability in capturing relevant quality information. As in \cite{agarla2021efficient}, the model is trained using the images from the CID \cite{virtanen2014cid2013} and the SPAQ \cite{fang2020perceptual} datasets with the aim of predicting the scores related to quality attributes, \textit{i.e.}, sharpness, groundness, lightness, saturation, brightness, colorfulness, noisiness, contrast, and MOS. In order to obtain the quality features for a video, 32 frames (the same used by the aesthetic encoder) are selected and processed by the network, and the features obtained before the FC layers of each branch are used. The resulting feature vector is obtained by concatenating the feature vectors produced by each branch. The feature combination module reduces the dimensions of the extracted features and prepares them to be used for quality score prediction. The quality-related attribute encoder extracts quality features frame-by-frame. Therefore, a GRU module is used to map frame-level quality features into a single vector capturing temporal dependency among them, and its dimension is later reduced by a FC layer. Then, these features are concatenated with the aesthetic features and the technical quality features related to the first half of the video, and processed by an additional FC layer. The same happens for the features obtained considering the technical features related to the second half of the video. The two outputs are later concatenated. Finally, in the quality prediction module the obtained feature vector is mapped into the final video quality score through an SVR.

There are about $78$ million parameters in the DOVER architecture and $20$ million parameters in the quality-related attribute encoder. Pretrained-models are utilized. Specifically, Video Swin Transformer is pretrained on LSVQ \cite{ying2021patch}, inflated-ConvNext is pretrained on the AVA dataset \cite{murray2012ava}, and EfficientNet-v2 is pretrained on the combination of CID \cite{virtanen2014cid2013} and SPAQ \cite{fang2020perceptual}. They not only utilize the provided VDPVE \cite{gao2023vdpve} dataset to train the whole model, but also take the CID and the SPAQ datasets as supplements. In the training phase, andom video fragments are used at training time, while five-crop video fragments are used at inference time. Video fragments are generated as described in \cite{wu2022disentangling}. The batch size is set to 5, the learning rate is set to $10^{-4}$ for the technical quality encoder and $10^{-3}$ for the rest of the network with a cosine decay. The model is trained for a total of 50 epochs. The quality-related attribute encoder is trained using the CID and SPAQ datasets. Images are first randomly cropped to the closest resolution that is multiple of 720p, and then soft pooling is applied to obtain a 720p resolution. The batch size is set to 8. The model is trained for a total of 10K iterations. The learning rate is initially set to $10^{-4}$ and later decreased by a factor of 10 after 5K iterations. Random horizontal flip is used as data augmentation. Both the encoders are trained using PLCC loss and rank loss, using the ground-truth scores as target. The rank loss has a weight of 0.3 in the total loss for training stability. The SVR for the final score prediction is trained on the VDPVE training set. The required hyperparameters, \textit{i.e.}, $\gamma = 12.20$ and $C = 364.83$, are selected via Bayesian optimization using Leave-One-Out cross-validation. In the testing phase, they average the quality scores obtained for the original video, and its horizontally flipped version.

\subsection{HXHHXH}
Team HXHHXH proposes a new pre-training method. To extract better spatial distortion features, they use the Live in the wild \cite{ghadiyaram2015massive} dataset to pre train the feature extraction network. Additionally, considering combination of temporal and spatial features can better achieve accurate quality assessment, they design a new temporal pre-training strategy. Concretely, for a video with length $L$, they sample new video segments of $L/2, L/4$ and $L/8$, and use these different video segments for pre-training to achieve better results. 

The whole model architecture is based on FastVQA \cite{wu2022fast}, along with a VGG network pre-trained using ImageNet as the feature extraction sub-model. The number of parameters is around $3$ million. For training details, they use Adam optimizer setting $\beta_1=0.9, \beta_2=0.999$. They set minibatch size as $12$, and learning rate as $0.0002$. The training process takes around $0.5$ hour on Nvidia GeForce RTX 3090.

\subsection{fmgtv}
The method proposed by team fmgtv can be divided into three parts: image quality assessment (IQA) module, VQA module, and fusion module. For IQA module, they extract each frame of the video and use a picture classification network for regression training, while the VQA module extracts $N$ frames at each interval and uses a video classification network for regression training. Specifically, they use ResNet-101, Convnext pre-trained using ImageNet as the feature extraction sub-model, Swin3D and Xclip pre-trained using LSVQ \cite{ying2021patch} dataset as backbone. Finally, they directly fuse the IQA and VQA results as their final resutls. The fusion strategy brings $3\%$ improvement ($0.74$ to $0.77$). 

During training, they use LSVQ \cite{ying2021patch} training dataset as extra data in addition to provided NTIRE 2023 training data. They use cosine learning rate initialized as $0.01$, and set batch size as $32$. The training process takes around twelve hours for $30$ epochs on Nvidia GeForce RTX 3090.

\subsection{KK-ARC}

Team KK-ARC proposes a method~\cite{huang2023sbvqa} by stacking ensemble on three VQA models: FastVQA-B \cite{wu2022fast}, FastVQA-M \cite{wu2022fast}, and FasterVQA \cite{wu2022neighbourhood}. All three models use Swin-Transformer \cite{liu2022video} as backbone, and are pre-trained on LSVQ \cite{ying2021patch} dataset. The provided NTIRE 2023 training dataset is used for finetuning on those models separately. XGBoost is then applied for stacking ensemble.  

\subsection{DTVQA}

\begin{figure*}[ht]
\centering
\includegraphics[width=\textwidth]{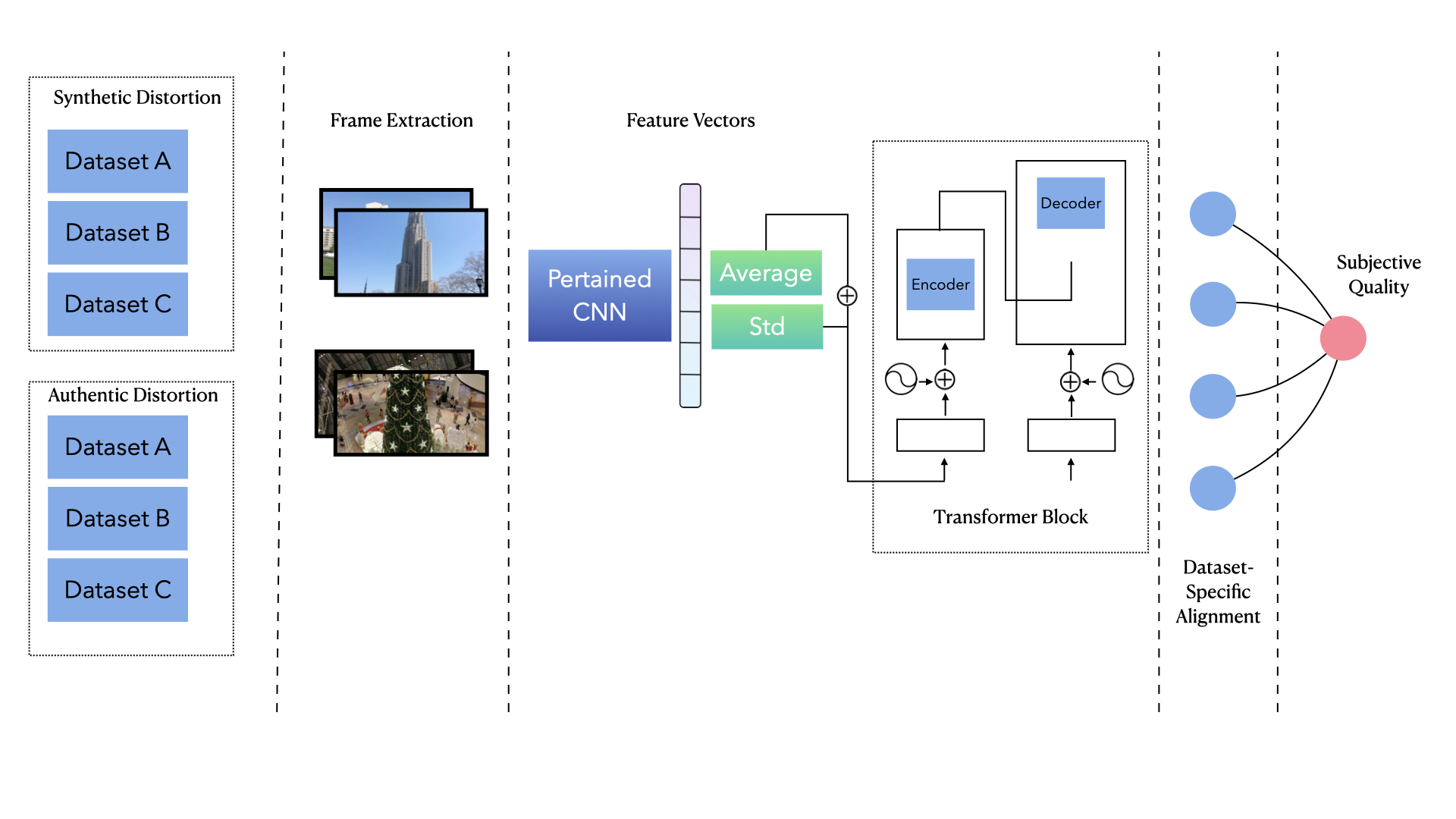}
\caption{The framework design of DTVQA team's method.}
\label{fig:dtvqa}
\end{figure*}

Team MTVQA proposes a self-attention based perception VQA method. The overview of their framework is shown in \ref{fig:dtvqa}. At the beginning, they divide VQA problem into authentic/aesthetic video quality and synthetic distortion VQA. After extracting frames from both parts, they use ResNet which is pre-trained using ImageNet as feature extraction sub-model. Different stages of feature map are extracted and concatenated together. They then implement dimensionality reduction by averaging the feature map. At the last step, a self-attention module is used to model the time dimension video quality.

Furthermore, they find that by adding more public available VQA datasets during training can overcome over-fitting. Specially, they combine two multi-dataset training strategy: 1) method proposed by MDTVSFA \cite{li2021unified}, which described a dataset-specific alignment method for training different datasets; 2) a multi-stage training strategy. After training the model on dataset A, they load the trained checkpoint and finetune the model on dataset B, and so on. The additional public available VQA datasets that are used are KoNVid-1k \cite{hosu2017konstanz}, YouTube-UGC \cite{wang2019youtube}, and MSU CVQA dataset \cite{antsiferova2022video}. 

The number of parameters for their model achieves approximately $1.71$ million. During training, the model is trained for 50 epochs using Adam optimizer with an initial learning rate $10^{-4}$. The batch size is set by 256 for each dataset. Whole training and testing processes are conducted on 4 Nvidia Tesla v100. It takes six hours for training, and 16ms per image during testing. 

It is worth mentioning that, to further overcome over-fitting, they conduct ensemble by averaging the outputs of the proposed method and DOVER \cite{wu2022disentangling} to get the final result.

\subsection{sqiyx}

Team members from sqiyx indicate that the provided data may not be enough to train a VQA model. As a result, they turn to an IQA model, MANIQA \cite{yang2022maniqa}. They introduce the fragment technology in FastVQA \cite{wu2022fast} to their method, which can keep the resolution information of the image from being lost, and can be regarded as a kind of data augmentation. To further augment data, they utilize cutMix to fuse two random images from the same video. Last but no least, they use a more large model than MANIQA \cite{yang2022maniqa} to increase the model capability. The feature extraction sub-model is a Vit Large Patch16 Network pre-trained using ImageNet. 

During training, they use Adam optimizer by setting $\beta_1 = 0.9$ and $\beta_2 = 0.999$. Minibatch size is set as $8$, and learning rate is initialized as $10^{-5}$. They use cosine scheduler to update learning rate with parameters Tmax and etamin set to 50 and 0. It takes approximately one day for training. During testing, they use pyav to extract key frames and calculate the results of each frame, and finally take the average to get the final score of the video. All experiments are conducted on one Nvidia Tesla v100.

\subsection{402Lab}

The method proposed by team 402Lab includes four parts, the spatial feature extraction module, the motion feature extraction module, the spatial-motion features fusion module, and the quality regression module. This method takes the first frame per second as input, and does not require any cropping operations, which results in low complexity and integral image information. ResNetv2-50 \cite{he2016identity} pre-trained on ImageNet is utilized as the feature extraction backbone network. Contestants propose a two-branch feature fusion network strategy, which effectively combines the advantages of CNN network and Transformer network,and fully integrates local and global information. In addition, they propose a patch attention module to make quality assessment more focused on effective information. After motion features and spatial features are extracted, we fuse
the two features using the spatial-motion feature fusion module, which is used to compensate for temporal-related distortions that cannot be modeled by spatial features.

The whole training is divided into two procedures. The model is first pre-trained on KonIQ-10K \cite{hosu2020koniq} and later finetuned on NTIRE 2023 training dataset. During pre-training, the initial learning rate of the backbone network is $10^{-5}$ and the rest is $10^{-4}$. They also randomly horizontally flipped images with a given probability of 0.5 during pre-training. During finetuning process, they use the SlowFast R50 \cite{feichtenhofer2019slowfast} as the motion feature extraction model for the whole experiments. The weights of the SlowFast R50 \cite{feichtenhofer2019slowfast} are fixed by training on the Kinetics 400 \cite{kay2017kinetics} dataset. The initial learning rate of the network is $10^{-5}$, and is reduced by $10$ after 80 epochs. AdamW optimizer is used in both pre-training and finetuning processes by setting $\beta_1 = 0.9, \beta_2 = 0.999$. And minibatch size is set as $8$. Furthermore, to maintain the image ratio, they first resize the image to $640 \times 360$, after which we use the same preprocessing method as in \cite{ying2020patches} to fill the image to a resolution of $640 \times 384$.
The same operation is conducted in testing process. 


\subsection{one\_for\_all}
This team proposes a VQA method based on the multi-clips ensemble, which contains
two steps: data filtering and partitioning based on video embedding clustering; and
quality content decoupled regression headers.

Their network contains about $55$ million parameters. They only use the training set of the VDPVE to train their network. No additional data has been used. They use the DOVER backbone pre-trained using LSVQ as the feature extraction sub-model. For optimization,they use the AdamW optimizer by setting $\beta_1$=$0.9$, $\beta_2$=$0.999$. They set the minibatch size as $8$. The learning rate is initialized as $10^{-3}$.

\subsection{NTU-SLab}

Team NTU-SLab proposes a network based on the DOVER, which consists of an aesthetic branch and a technical branch. Moreover, they ensemble the DOVER result with raw feature tuning from CLIP-RN50 visual backbone. 

Their network contains around $75$ million parameters. They only use the training set of the VDPVE to train their network. No additional data has been used. For optimization, they use Adam optimizer by setting $\beta_1$=$0.9$, $\beta_2$=$0.999$. They set minibatch size as 8 and train for 30 epochs. The learning rate is initialized as $10^{-3}$ and kept unchanged during training. They ensemble the DOVER and CLIP-RN50
results with 2:1 ratio.

\subsection{HNU-LIMMC}
In order to improve the sensitivity of the model to enhanced video perception, they propose a novel contrastive learning method for VQA based on the idea of self-supervision to improve the performance of the model. At the same time, they introduce a video degradation space. Specifically, they believe that different frames of the same video should be similar and can effectively represent the video. The quality of different degradation methods of the same frame is not similar. 

Through this framework, an end-to-end learning method can be effectively established. The degraded video information in the VQA dataset is used to simulate the learning method in the enhanced scene, which is conducive to the generalization ability of the model and improves the effect of the model on the non-natural VQA dataset.

Their network contains about 2.472 million parameters. They only use the training set of the VDPVE to train their network. No additional data has been used. They choose SimpleVQA as the backbone of their model, which is pre-trained on LSVQ and fine-tuned on VDVPE. For optimization, they use Adam optimizer. They set the minibatch size as 6. The learning rate is initialized as $10^{-5}$ and the weightdecay is $10^{-8}$. 
\subsection{Drealitym}
They propose a NR VQA method for enhanced videos based on the framework of Adaptive Token-Selection ViT (ATSViT) that the process of energy competition between visual information from the psychological perspective to predict video quality scores.
They propose a block-level sampling strategy, called timing block sampling (TBS), that takes into account the uneven distribution of local quality distortions focused on by the human eye in the original sequence, increasing the information density of the sampled frame set and reducing the loss possibility of important spatial features through HVS based \cite{theeuwes1998our,liu2021video} fine-grained sampling. They construct transformer-based Stage-wise adaptive Screening Network (SSNet) based on based on the filter theory of attention \cite{wood1995cocktail}, dividing the process of visual information processing into four stages where efficient energy distribution strategies are used, exploiting the attention-based bottlenecks of different sizes, to select the features of tokens that advance to the next stage.

They only use the training set of the VDPVE to train their network. No additional data has been used. They use swin-B Network pretrained on the Kinetics-400 dataset to initialize the backbone in SSNet. For optimization, they use the AdamW optimizer by setting $\beta_1$ = 0.9, $\beta_2$ = 0.999. They set minibatch size as 5. The learning rate is initialized as $5\times10^{-6}$ and use the custom warmup cosine step decay to updates the learning rate. The Weight decay is set as $1.5\times10^{-3}$.
\subsection{LION\_Vaader}
\begin{figure*}[h!]
\subfloat[Illustration of the features extraction process]{\includegraphics[width=\textwidth]{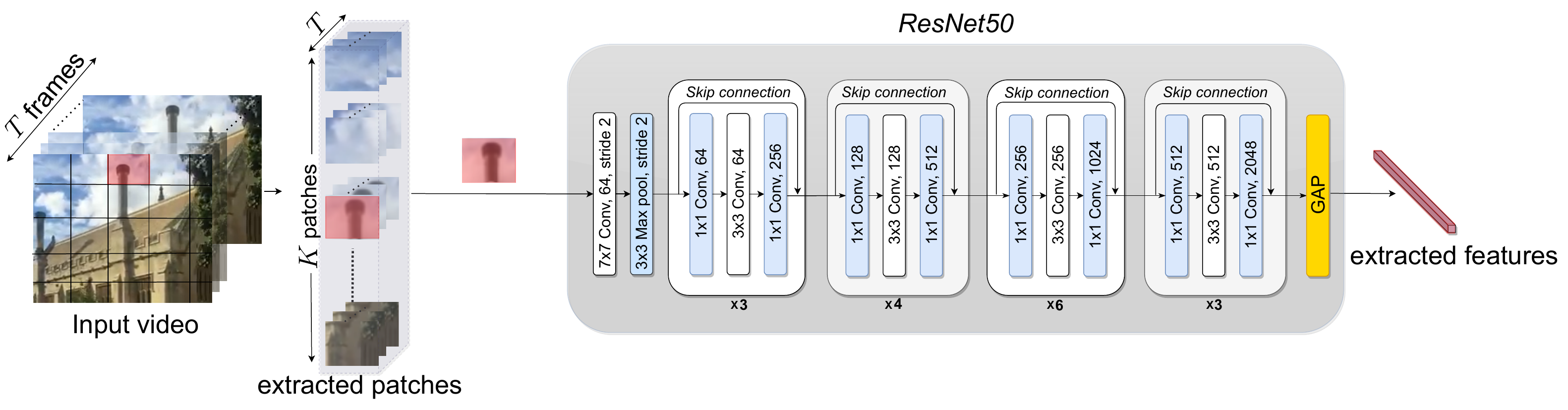}}\label{fig:resnet50}
\hfil
\subfloat[Illustration of the feature pooling process]{\includegraphics[width=\textwidth]{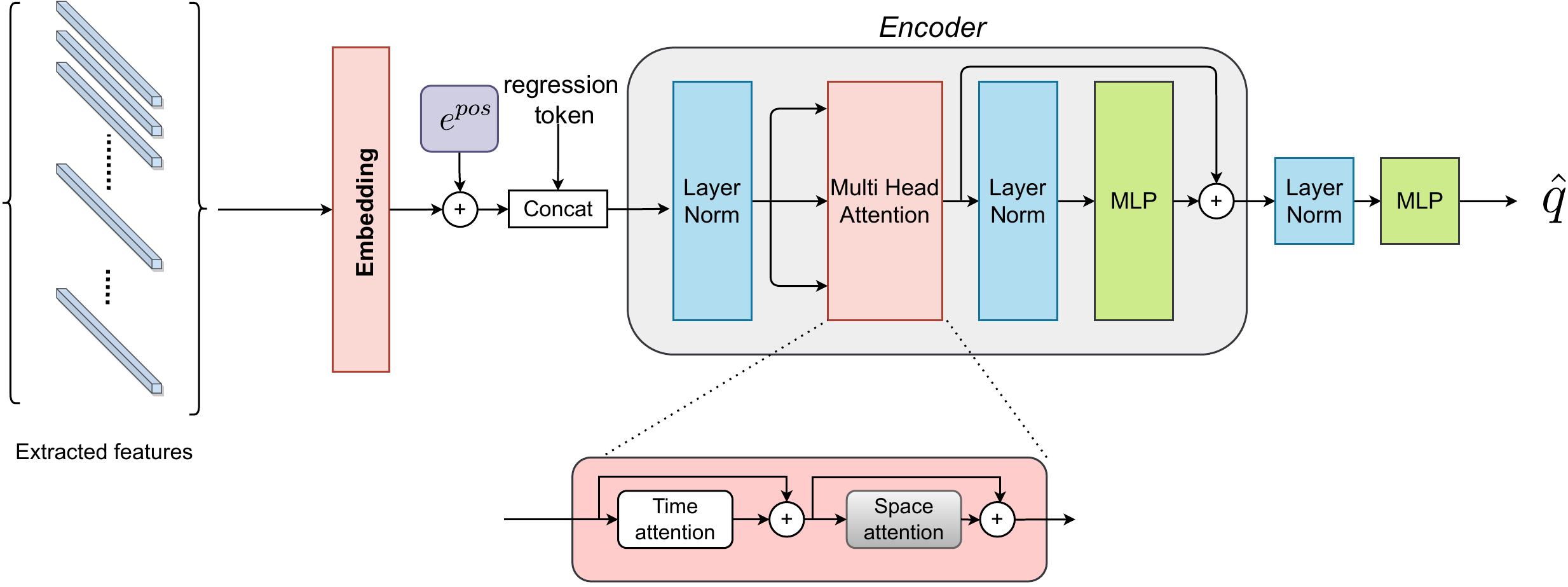}\label{fig:timeSformer}}
\hfil
\caption{The framework design of LION\_Vaader team's method.}
\label{fig:proposed}
\end{figure*}
They adopt the well-known ResNet-50 model and exploit it to perform transfer learning via a fine-tuning technique using the weights of the ImageNet dataset.
First, they perform a temporal sub-sampling of the video by selecting one
frame per second. Then, from each of the selected frames, a set of
patches is extracted in a sliding window fashion so that their dimensions match the standard architecture's input shape.
Second, each patch is fed to the CNN backbone for feature extraction and fits into the spatio-temporal pooling module in a time-distributed fashion.
The extracted features are then fed to the spatio-temporal pooling module. The use of this pooling type is motivated by the fact that the visibility of an artifact depends highly on its location and its neighboring regions in the current frame and the adjacent frames, resulting in the so-called masking phenomenon. Thus, the overall quality of a given video is affected by both, spatial artifacts that occur in some regions of the frame, as well as temporal artifacts that affect a range of sequential frames. 

Most of the pooling techniques are effective at capturing short-range patterns within local spatio-temporal regions, whereas they can only  model space-time dependencies of at most a handful of seconds, not video whole. To address that, they use a spatio-temporal transformer namely TimeSformer as pooling architecture, which exploits space-time attention. Thus, the feature pooling module captures short-term dependencies between neighboring patches as well as long-range correlations between distant patches. In addition, this pooling can analyze the video over much longer time spans.

Their network contains $25.6$ million parameters in the CNN resnet50 model and $122.13$ million parameters in the TimeSformer architecture (vary according to the input’s number of frames and patches). They use the All Combined dataset that the authors in \cite{tu2021ugc} proposed by merging the KoNViD-1k, LIVE-VQC and YouTube-UGC datasets. They use the ResNet50 pre-trained on ImageNet as the feature extraction sub-model. For optimization, they use the Adam optimizer with default parameters. They set minibatch size as $1$ due to the diversity of the number of frames and patches that can be extracted per video. The learning rate is initialized as $10^{-3}$, and a reduce learning rate on plateau call back is used with patience parameter set to $5$. Features are only extracted once, and used directly to train the transformer architecture. 

\subsection{Caption Timor}
This team utilized SimpleVQA to directly extract spatial and mobile features, while applying random rotation, flipping, and cutting as data augmentation techniques. They optimized the model by adjusting parameters and taking the average value of five models obtained from five-fold cross-validation training. The loss function used was MSE loss.

Their network consists of $2.6$ million parameters and was trained solely on the VDPVE training set without additional data. They employed a ResNet50 pre-trained on ImageNet as the feature extraction sub-model. The Adam optimizer was used for optimization, with $\beta_1$ set to 0.9 and $\beta_2$ set to 0.999. The minibatch size was set at 8, with a learning rate initialized at $10^{-5}$ and halved every $5\times10^4$ minibatch updates. During the testing phase, the batch size was set to 1.

\subsection{IVP-LAB}
They introduce a novel method to acquire frame level deep features for assessing the quality of videos. To accomplish this, they focus on the deep feature maps correlations of a pre-trained network, or more specifically, their similarity as a helpful tool for assessing video quality. The covariance matrix \textit{i.e.} the Gram matrix, which depicts the correlation between all feature maps of a specific mid-layer, can be stated as deep feature relationships. The structural details of frames’ appearance are reflected in these relations and significantly correlate with the perceived quality of a given video. The extracted feature maps relations in different granularities can effectively illustrate the influence of various distortions.
Every feature map reflects a different structural detail of the source image. It is shown in \cite{gatys2016image} that almost flawless reconstruction is possible from the network’s lower layers whereas the detailed pixel information is insufficiently maintained in the network’s upper layers. In this case, each layer’s output of a convolutional neural network can be represented by a collection of feature maps that show the input pixels’ structural data. In the proposed method, the extracted Gram Matrix of mid-level convolutional layer is employed as the frame level feature. The proposed method exploits the correlation between the deep feature maps derived from each network’s layers to assess the video’s quality.

Their network contains $21.7855$ million parameters. They only use the training set of the VDPVE to train their network. No additional data has been used. They use the inception-v3 pre-trained on ImageNet as the feature extraction sub-model. The only trainable model is a linear SVR model that is trained using features extracted by a pretrained inception-v3
model. They set epsilon value as $0.3$. The first test phase involves extracting spatial information from video frames and merging it into a feature vector. Then this feature vector was given to an SVR model to predict the quality score.

\section*{Acknowledgments}

We thank Peng Cheng Laboratory for sponsoring this NTIRE 2023 challenge and the NTIRE 2023 sponsors: Sony Interactive Entertainment, Meta Reality Labs, ModelScope, ETH Z\"urich (Computer Vision Lab) and University of W\"urzburg (Computer Vision Lab).

\appendix

\section{NTIRE 2023 Organizers}
\noindent\textit{\textbf{Title: }}\\ NTIRE 2023 Quality Assessment of Video Enhancement Challenge\\
\noindent\textit{\textbf{Members:}}\\ 
 \textit{Xiaohong Liu$^1$ (xiaohongliu@sjtu.edu.cn)}, Xiongkuo Min$^1$, Wei Sun$^1$,  Yulun Zhang$^2$, Kai Zhang$^2$,  Radu Timofte$^{2,3}$, Guangtao Zhai$^1$,  Yixuan Gao$^1$,  Yuqin Cao$^1$, Tengchuan Kou$^1$,  Yunlong Dong$^1$, Ziheng Jia$^1$\\
\noindent\textit{\textbf{Affiliations: }}\\
$^{1}$ Shanghai Jiao Tong University, China\\
$^2$ ETH Z\"urich, Switzerland\\
$^3$ University of W\"urzburg, Germany\\

\section{Teams and Affiliations}
\label{sec:apd:track1team}

\renewcommand{\thefootnote}{\fnsymbol{footnote}}
\subsection*{TB-VQA}
\noindent\textit{\textbf{Title:}}\\
Video Quality Assessment Based on Swin Transformer with Spatio-Temporal Feature Fusion and Data Augmentation\\
\noindent\textit{\textbf{Members: }}\\
\textit{Yilin Li$^1$ (gustav.lyl@alibaba-inc.com)}, Wei Wu$^1$, Shuming Hu$^1$, Sibin Deng$^1$, Pengxiang Xiao$^1$\footnote{Pengxiang Xiao is also with Vrobotit Lab, Beijing University of Posts and Telecommunications, and the work is primarily done during an internship at Alibaba Group.}, Ying Chen$^1$, Kai Li$^1$ \\
\noindent\textit{\textbf{Affiliations: }}\\
$^1$ Department of Tao Technology, Alibaba Group\\

\subsection*{QuoVadis}
\noindent\textit{\textbf{Title:}}\\
A Dual-branch Network for Enhanced Video Quality Assessment\\
\noindent\textit{\textbf{Members: }}\\
\textit{Kai Zhao$^1$ (zhaokai05@kuaishou.com)}, Kun Yuan$^1$, Ming Sun$^1$\\
\noindent\textit{\textbf{Affiliations: }}\\
$^1$ Kuaishou Technology\\

\subsection*{OPDAI}
\noindent\textit{\textbf{Title:}}\\
Apply Image Quality Assessment into Video Quality Assessment\\
\noindent\textit{\textbf{Members: }}\\
\textit{Heng Cong$^1$ (congheng@corp.netease.com)}, Hao Wang$^1$, Lingzhi Fu$^1$, Yusheng Zhang$^1$, Rongyu Zhang$^1$\\
\noindent\textit{\textbf{Affiliations: }}\\
$^1$ Interactive Entertainment Group of Netease Inc, Guangzhou, China\\

\subsection*{TIAT}
\noindent\textit{\textbf{Title:}}\\
A Deep Learning based Video Quality Assessment Model\\
\noindent\textit{\textbf{Members: }}\\
\textit{Hang Shi$^1$ (hang.shi@transsion.com)}, Qihang Xu$^1$, Longan Xiao$^1$\\
\noindent\textit{\textbf{Affiliations: }}\\
$^1$ Transsion\\

\subsection*{VCCIP}
\noindent\textit{\textbf{Title:}}\\
Channel Fusion for Video Quality Assessment\\
\noindent\textit{\textbf{Members: }}\\
\textit{Zhiliang Ma$^1$ (mzl@mail.hfut.edu.cn)}\\
\noindent\textit{\textbf{Affiliations: }}\\
$^1$ Hefei University of Technology\\

\subsection*{IVL}
\noindent\textit{\textbf{Title:}}\\
    Video Quality Assessment Guided by Aesthetics and Technical Quality Attributes\\
\noindent\textit{\textbf{Members: }}\\
\textit{Mirko Agarla$^1$ (m.agarla@campus.unimib.it)}, Luigi Celona$^1$, Claudio Rota$^1$, Raimondo Schettini$^1$\\
\noindent\textit{\textbf{Affiliations: }}\\
$^1$ Department of Informatics Systems and Communication, University of Milano - Bicocca\\

\subsection*{HXHHXH}
\noindent\textit{\textbf{Title:}}\\
    A Multi-interval Sampling Strategy Pre-training for Video Quality Assessment\\
\noindent\textit{\textbf{Members: }}\\
\textit{Zhiwei Huang$^1$ (huangzhiwei10@xiaomi.com)}, Ya’nan Li$^1$, Xiaotao Wang$^1$, Lei Lei$^1$\\
\noindent\textit{\textbf{Affiliations: }}\\
$^1$ Xiaomi Inc., China\\

\subsection*{fmgtv}
\noindent\textit{\textbf{Title:}}\\
    Integration of IQA and VQA\\
\noindent\textit{\textbf{Members: }}\\
\textit{Hongye Liu$^1$ (liuhongye1998@163.com)}, Wei Hong$^2$\\
\noindent\textit{\textbf{Affiliations: }}\\
$^1$ China Ji Liang University\\
$^2$ FreeTech\\

\subsection*{KK-ARC}
\noindent\textit{\textbf{Title:}}\\
    Stacking Ensemble with FastVQA-B, FastVQA-M, and FasterVQA\\
\noindent\textit{\textbf{Members: }}\\
\textit{Ironhead Chuang$^1$ (ironheadchuang@kkcompany.com)}, Allen Lin$^1$, Drake Guan$^1$, Iris Chen$^1$, Kae Lou$^1$, Willy Huang$^1$, Yachun Tasi$^1$, Yvonne Kao$^1$\\
\noindent\textit{\textbf{Affiliations: }}\\
$^1$ Advanced Research Center, KKCompany, Taiwan\\

\subsection*{DTVQA}
\noindent\textit{\textbf{Title:}}\\
    Self-Attention based Perception Video Quality Assessment Method\\
\noindent\textit{\textbf{Members: }}\\
\textit{Haotian Fan$^1$ (fanhaotian@bytedance.com)}, Fangyuan Kong$^1$\\
\noindent\textit{\textbf{Affiliations: }}\\
$^1$ ByteDance\\

\subsection*{sqiyx}
\noindent\textit{\textbf{Title:}}\\
    No Title\\
\noindent\textit{\textbf{Members: }}\\
\textit{Shiqi Zhou$^1$ (408172566@qq.com)}, Hao Liu$^1$\\
\noindent\textit{\textbf{Affiliations: }}\\
$^1$ MGTV\\

\subsection*{402Lab}
\noindent\textit{\textbf{Title:}}\\
    Triple-Branch Feature Fusion Network, Spatial Feature Extraction Subnetwork, Spatial-Motion Features Fusion Mdule\\
\noindent\textit{\textbf{Members: }}\\
\textit{Yu Lai$^1$ (2672339375@qq.com)}, Shanshan Chen$^1$\\
\noindent\textit{\textbf{Affiliations: }}\\
$^1$ Fuzhou University\\

\subsection*{one\_for\_all}
\noindent\textit{\textbf{Title:}}\\
Video Quality Assessment based on Multi-Clips Ensemble
\noindent\textit{\textbf{Members:}}\\
\textit{Wenqi Wang$^1$(wwenki1992@163.com)},\\
\noindent\textit{\textbf{Affiliations: }}\\
$^1$ Shopee Information Technology Co., Ltd.\\
\subsection*{NTU-SLab}
\noindent\textit{\textbf{Title:}}\\
DOVER-CLIP-RN50\\
\noindent\textit{\textbf{Members: }}\\
\textit{Haoning Wu$^1$(haoning001@e.ntu.edu.sg)}, {Chaofeng Chen$^1$ }\\
\noindent\textit{\textbf{Affiliations: }}\\
$^1$ S-Lab, Nanyang Technological University\\
\subsection*{HNU-LIMMC}
\noindent\textit{\textbf{Title:}}\\
Exploring Comparative Learning-Inspired Strategies for Video Quality Evaluation for Enhance Video\\
\noindent\textit{\textbf{Members: }}\\
\textit{Chunzheng Zhu$^1$ (1724735214@qq.com)}, {Zekun Guo$^1$ }
\noindent\textit{\textbf{Affiliations: }}\\
$^1$ Hunan University\\
\subsection*{Drealitym}
\noindent\textit{\textbf{Title:}}\\
Video Transformer based Video Quality Assessment\\
\noindent\textit{\textbf{Members: }}\\
\textit{Shiling Zhao$^1$ (yiyiaiou@163.com)}, {Haibing Yin$^1$}, {Hongkui Wang$^1$}\\
\noindent\textit{\textbf{Affiliations: }}\\
$^1$ Hangzhou Dianzi University\\
\subsection*{LION\_Vaader}
\noindent\textit{\textbf{Title:}}\\
Quality Assessment for Video Enhancement using Joint Space-Time Attention\\
\noindent\textit{\textbf{Members: }}\\
\textit{Hanene Brachemi Meftah$^1$(hanene.brachemi@insa-rennes.fr)}, {Sid Ahmed Fezza$^2$}, {Wassim Hamidouche$^1$}, {Olivier Déforges$^1$}\\
\noindent\textit{\textbf{Affiliations: }}\\
$^1$ INSA Rennes, CNRS, IETR - UMR 6164, Rennes, France\\
$^2$ National Higher School of Telecommunications and ICT, Oran, Algeria\\
\subsection*{Caption Timor}
\noindent\textit{\textbf{Title:}}\\
Five Fold and Data Augmentation.\\
\noindent\textit{\textbf{Members:}}\\
\textit{Tengfei Shi$^1$ (tengfeishibh@163.com)}\\
\noindent\textit{\textbf{Affiliations: }}\\
$^1$ BUAA\\
\subsection*{IVP-LAB}
\noindent\textit{\textbf{Title:}}\\
Feature Maps Correlation-based Video Quality Assessment
\noindent\textit{\textbf{Members: }}\\
\textit{Azadeh Mansouri$^1$(a\_mansouri@khu.ac.ir)},
{Hossein Motamednia$^2$},
{Amir Hossein Bakhtiari$^1$},
{Ahmad Mahmoudi Aznaveh$^3$}\\
\noindent\textit{\textbf{Affiliations: }}\\
$^1$ Department of Electrical and Computer Engineering Faculty of Engineering Kharazmi University, Tehran, Iran\\
$^2$ High Performance Computing Laboratory School of Computer Science Institute for Research in Fundamental Sciences
Tehran, Iran\\
$^3$ Cyberspace Research Institute Shahid Beheshti University, Tehran, Iran\\


{\small
\bibliographystyle{ieee_fullname}
\bibliography{egbib}
}

\end{document}